\documentclass[12pt]{article}

\usepackage[utf8]{inputenc}
\usepackage{bbm}
\usepackage{bm}
\usepackage{ragged2e}
\usepackage{mathrsfs}
\usepackage{amsfonts}
\usepackage{float}
\usepackage{amsthm}
\usepackage{epsfig,rotating}
\usepackage{amssymb,amsmath}
\usepackage{latexsym}
\usepackage{graphicx}
\usepackage{caption,subcaption}
\usepackage{multirow}
\usepackage{booktabs,natbib}
\usepackage{comment}
\usepackage{graphicx}
\graphicspath{{Graph/}}
\usepackage{xcolor}
\usepackage{diagbox}
\usepackage{comment}
\usepackage{tabularx}
\usepackage{algorithmicx}
\usepackage[ruled,vlined]{algorithm2e}

\usepackage{xurl}
\usepackage{geometry}
\usepackage{multibib}
\usepackage{endnotes}
\usepackage[titletoc]{appendix}
\bibpunct{(}{)}{;}{a}{,}{,}
\newcites{New}{References}

\newtheorem{definition}{\noindent D{\footnotesize EFINITION}}

\newtheorem{theorem}{\noindent T{\footnotesize HEOREM}}
\newtheorem{proposition}{\noindent P{\footnotesize ROPOSITION}}

\newtheorem{lemma}{\noindent L{\footnotesize EMMA}}
\newtheorem{coro}{ \noindent C{\footnotesize OROLLARY}}
\newtheorem{remark}{\noindent R{\footnotesize EMARK}}

\newcommand{\RN}[1]{%
  \textup{\uppercase\expandafter{\romannumeral#1}}%
}

\newcommand{\la}{\langle}
\newcommand{\ra}{\rangle}

\def\nano{\scriptscriptstyle}
\newcommand\hi[1]{^{#1}}
\def\real{\mathbb R}
\newcommand\ca[1]{{\cal{#1}}}

\newcommand\lo[1]{_{#1}}
\def\vec{\mbox{vec}}

\def\oc{\hi{\perp}}

\def\sign{\mathrm{sign}}

\def\trace{\mathrm{trace}}
\def\nano{\scriptscriptstyle}
\def\inv{\hi{\nano -1}}
\def\oc{\hi{\nano \perp}}

\def\nano{\scriptscriptstyle}

\def\diag{\mathrm{diag}}

\def\hii#1{\hi{(#1)}}

\def\diag{\mbox{diag}}

\newcommand{\trans}{^{\mbox{\tiny {\sf T}}}}
\newcommand{\indep}{\rotatebox[origin=c]{90}{$\models$}}
\def\var{\mathrm{var}}

\newcommand{\xdownarrow}[1]{%
  {\left\downarrow\vbox to #1{}\right.\kern-\nulldelimiterspace}
}

\DeclareMathOperator{\argmin}{\mathrm{argmin}}
\DeclareMathOperator{\argmax}{\mathrm{argmax}}
\makeatletter
\def\old@comma{,}
 \catcode`\,=13
 \def,{%
   \ifmmode%
     \old@comma\discretionary{}{}{}%
   \else%
     \old@comma%
   \fi%
 }
 \makeatother

\title{A Copula Graphical Model for Multi-Attribute Data using Optimal Transport}
\author{
\normalsize Qi Zhang, Bing Li, and Lingzhou Xue\\ \normalsize Department of Statistics, The Pennsylvania State University
}
\date{}

\begin{document}

\maketitle
\begin{abstract}
Motivated by modern data forms such as images and multi-view data, the multi-attribute graphical model aims to explore the conditional independence structure among vectors. Under the Gaussian assumption, the conditional independence between vectors is characterized by blockwise zeros in the precision matrix. To relax the restrictive Gaussian assumption, in this paper, we introduce a novel semiparametric multi-attribute graphical model based on a new copula named {Cyclically Monotone Copula}. This new copula treats the distribution of the node vectors as multivariate marginals and transforms them into Gaussian distributions based on the optimal transport theory. Since the model allows the node vectors to have arbitrary continuous distributions, it is more flexible than the classical Gaussian copula method that performs coordinatewise Gaussianization. We establish the concentration inequalities of the estimated covariance matrices and provide sufficient conditions for selection consistency of the group graphical lasso estimator. For the setting with high-dimensional attributes, a {Projected Cyclically Monotone Copula} model is proposed to address the curse of dimensionality issue that arises from solving high-dimensional optimal transport problems. Numerical results based on synthetic and real data show the efficiency and flexibility of our methods.
\end{abstract}



\section{Introduction}
The undirected graphical model is commonly used to study the conditional independence relations among a set of random variables. Given a $p$-dimensional random vector $X$, the goal is to estimate the undirected graph $\ca G = (V, E)$, where $V$ is the node set of cardinality $p$, corresponding to the indices components of $X$, $E$ is the edge set indicating whether two nodes are connected, which happens if and only if the two random variables on the nodes are conditionally dependent. For multivariate Gaussian $X \sim N(\mu, \Sigma)$, the precision matrix $\Theta=\Sigma^{-1}$ fully characterizes the conditional independence relations, making sparse precision matrix estimation the main approach for high-dimensional graph estimation.  Methods for this include the neighborhood selection \citep{meinshausen2006high}, Graphical Lasso \citep{yuan2007model,friedman2008sparse}, constrained $\ell_1$-minimization \citep{cai2011constrained}, and penalized D-trace loss \citep{zhang2014sparse,tao2021additive}. To relax the Gaussian assumption, \citet{liu2009nonparanormal} proposed a Copula Gaussian Graphical Model (Copula-GGM), which assumes that $p$ marginal transformations can transform the data to multivariate Gaussian. The model coincides with the Gaussian copula when the transformations are monotone. Rank-based correlation estimators for the Copula-GGM are proposed by \citet{liu2012high} and \citet{xue2012regularized}. A comprehensive review of this direction is available in \cite{lafferty2012sparse}. The advantage of the copula approach is it retains the simplicity of the conditional independence structure while allowing the marginal distribution to be arbitrary.

The classical undirected graphical model only considers scalar random variables on nodes. However, in modern applications, there is a need for a graphical model with nodes representing multi-attribute entities or random vectors. \citet{kolar2013markov,kolar2014graph} developed the multi-attribute graphical model where nodes correspond to vectors representing multi-attribute entities and edges encode the conditional dependence between vectors. The model has been successfully applied in various domains, including uncovering gene regulatory networks from gene and protein profiles \citep{chiquet2019multiattribute}, inferring the brain connectivity graph from positron emission tomography data \citep{kolar2014graph}, and inferring color image graphs by modeling the dependence between pixels \citep{tugnait2021sparse}.

\begin{figure}[ht!]
    \centering
    \includegraphics[width=\textwidth]{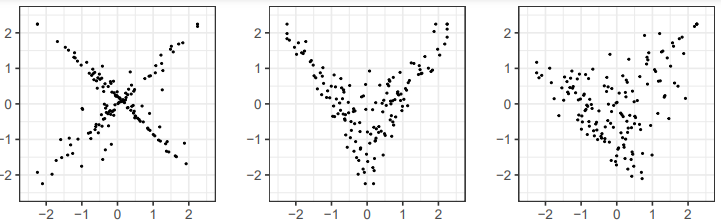}
    \caption{Visualization of jointly non-Gaussianity among 2-dimensional data}
    \label{fig: non_gauss}
\end{figure}

In the multi-attribute graphical model, when the data follows a multivariate Gaussian distribution, conditional independence can still be inferred from the corresponding block in the precision matrix. To relax the Gaussian assumption, we may still apply a coordinate-wise monotone transformation as we did for classical Gaussian graphical model. However, the coordinatewise Gaussianization is unnecessarily strong in the multi-attribute setting. To see the situation clearly, consider the following assumptions:
\begin{enumerate}
\item after transforming every element of $X$ to Gaussian, $X$ is jointly Gaussian;
\vspace{-0.11in}
\item after transforming every node vector of $X$ to Gaussian, $X$ is jointly Gaussian.
\end{enumerate}
By logic, the first assumption implies the second. The first statement is what underlies the current copula Gaussian graphical models;   the second uderlies the the new copula model we propose. To provide more intuition, in Figure \ref{fig: non_gauss} we plot three scenarios: a cross-shaped distribution, a v-shaped distribution, and a triangle-shaped distribution. In all cases, the two elements are marginally distributed as Gaussian, but jointly the vectors are  strongly non-Gaussian. If we have three node vectors having these three distributions, then no marginal transformation can lead to joint Gaussian distribution $X$.
This motivates us to develop more flexible semiparametric models that link multi-dimensional marginals in a graph.

More rigorously, a multi-attribute graphical model can be viewed as a random vector $X=(X\lo 1\trans, X\lo 2\trans,\dots, X\lo p\trans)\trans \in \oplus_{j=1}^p\real\hi {d\lo j}$ along with a graph $\ca G$ of $p$ nodes, where nodes $j$ and $k$ are connected if and only if $X\lo j$ and $X\lo k$ are conditionally dependent given $X\lo{-(j,k)} = \{X\lo \ell, \ell\neq j, k\}$. To link the multivariate marginals, we introduce a copula, called the {Cyclically Monotone Copula}, based on optimal transport theory. Specifically, we solve an optimal transport problem from the distribution of each node vector to a Gaussian distribution, and assume that optimal transport maps $\{T\lo j\}\lo {j=1}\hi p$ transform the entire $X$ to joint Gaussian. The name Cyclically Monotone describes the geometric structure of the optimal transport map based on the result of \citet{brenier1991polar}. This copula model is very flexible, as it allows the multivariate marginals to be arbitrary continuous distributions, including the scenario described in Figure \ref{fig: non_gauss}.

Our approach is inspired by the multivariate ranks introduced by \cite{chernozhukov2017monge}, which have been used for multivariate independence test \citep{ghosal2019multivariate,deb2021multivariate,shi2022distribution}, vector quantile regression \citep{carlier2016vector}, among others. Two related works include the  semiparametric CCA model developed by \cite{bryan2021multirank} based on a generalized Gaussian copula, and the vector copula based on measure transportation proposed in \cite{FAN2022}. In this paper, we use the proposed copula to estimate high-dimensional graphical models.

Estimating the cyclically monotone transformations $\{T\lo j\}\lo{j=1}\hi p$ directly requires solving discrete-to-discrete optimal transport problems, which suffers from the curse of dimensionality issue with a minimax rate of $O(n\hi{-1/d})$  \citep{fournier2015rate,niles2022estimation}. This can result in an inaccurate estimation of the graph structure, especially when the numbers of attributes on some nodes are large. To address this, we propose a  {projected cyclically monotone copula} (PCMC) for the multi-attribute graphical model with large-dimensional attributes. We assume that the non-Gaussianity of the $d$-dimensional attributes only appears in a low-dimensional subspace. After properly estimate the low-dimensional non-Gaussian subspace and perform Gaussianization within it, the data can be efficiently transformed into a Gaussian distribution. This idea stems from the recent developments of projection-based techniques for solving high-dimensional optimal transport problems, such as the slicing approach \citep{deshpande2018generative}, projection pursuit approach \citep{meng2019large}, and projection robust method \citep{paty2019subspace}. A comprehensive review of this direction is provided in \citet{zhang2022projection}. 

The rest of the paper is organized as follows. In Section \ref{sec: CMC-GGM}, we introduce the cyclically monotone copula Gaussian graphical model (CMC-GGM) for multi-attribute data and extend it to the composition cyclically monotone copula Gaussian graphical model (CCMC-GGM). Estimation methods are developed, and their consistency and convergence rates established, in Sections \ref{sec: estimation} and \ref{sec: asymptotic}. In Section \ref{sec: pcmc}, we propose the projected cyclically monotone copula Gaussian graphical Model (PCMC-GGM) to avoid the curse of dimensionality arising from solving optimal transport problems. In Section \ref{sec: simulation},  we present simulation results that compare our methods with Gaussian graphical models and copula Gaussian graphical models for multi-attributes data. Finally, in Section \ref{sec: application}, we apply the CMC-GGM to estimate the gene and protein regulatory network and color texture graph.

\section{Background}
\subsection{Multi-Attribute Graphical Model}\label{mggm}
Consider $p$ random vectors $X\lo j \in \real\hi {d\lo j}$ for $j=1,\dots, p$. We assume $d=d\lo1=\cdots = d\lo p$ for notational simplicity, but the model can be easily extended to cover distinct $d$'s. We let $X=(X\lo 1\trans,X\lo 2\trans,\dots,X\lo p\trans)\trans\in\real\hi {dp}$. Let $\ca G(V, E)$ be the undirected graph with node set $V=\{1,2,\dots,p\}$ and edge set $E\subseteq\{(j,k)\in V\times V: j\neq k\}$. 
We say $X$ follows a multi-attribute graphical model with the graph $\ca G(V, E)$ if the edge set $E$ is determined by the conditional independence structure among $X\lo1,\dots, X\lo p$ in the following way: 
\begin{equation}\label{eq: graph_equi}
    X_j\indep X_k|X_{-(j,k)}\,\, \text{if and only if}\,\, (j,k)\notin E,
\end{equation}
where $X\lo{-(j,k)} = \{X\lo \ell, \ell \neq j, k\}$. We call $X\lo1, \dots, X\lo p$ the node vectors. If, furthermore, $X$ follows a multivariate Gaussian distribution $N(0,\Sigma)$, where $\Sigma$ and $\Theta =\Sigma\inv$ have block structures
\begin{equation}\label{eq: cov_pre}
 \Sigma = \begin{pmatrix}
    \Sigma\lo{11}& \Sigma\lo{12}&\dots&\Sigma\lo{1p}\\
    \Sigma\lo{21}& \Sigma\lo{22}&\dots&\Sigma\lo{2p}\\
    \vdots& \vdots & \ddots& \vdots\\
    \Sigma\lo{p1} & \Sigma\lo{p2}& \cdots & \Sigma\lo{pp}
    \end{pmatrix}\,\, \text{and}\,\,
    \Theta=\begin{pmatrix}
    \Theta\lo{11}& \Theta\lo{12}&\dots&\Theta\lo{1p}\\
    \Theta\lo{21}& \Theta\lo{22}&\dots&\Theta\lo{2p}\\
    \vdots& \vdots & \ddots& \vdots\\
    \Theta\lo{p1} & \Theta\lo{p2}& \cdots & \Theta\lo{pp}
    \end{pmatrix},
\end{equation}
then we refer to the model as the Gaussian Graphical Model (GGM) for multi-attributes data. Under the Gaussian assumption, the conditional independence in \eqref{eq: graph_equi} can be replaced by $\Theta\lo {jk}=0$; that is, there is an edge between $j$ and $k$ iff $\Theta\lo {jk} \neq 0$.

Compared with the single-attribute graphical model, the multi-attribute graphical model has a dependence structure within each node, parametrized by sub-precision matrices $\Theta\lo {jj}, j=1, \dots, p$ on the diagonal of $\Theta$. However, to infer the edge set $E$ of the graph, we only need to consider the conditional dependence relations across different node vectors. Therefore, any one-to-one transformations on $\{X\lo j\}\lo{j=1}\hi p$ that preserves the conditional dependence structure can be applied. For instance, there exist orthogonal matrices $\Gamma\lo j\in \real\hi {d \times d}$ such that $\Gamma\lo j X\lo j\sim N(0, I\lo d)$ for $j =1 , \dots, p$.  Let $\Gamma = (\Gamma\lo 1,\dots, \Gamma\lo p)$ and $\tilde X = \Gamma X$. The graph estimated using $X$ is equivalent to that of $\tilde X$. Therefore, without loss of generality, we can always assume the diagonal blocks $\{\Sigma\lo {jj}\}\lo{j=1}\hi p$ to be the identity matrices.

\subsection{Optimal Transport and Cyclical Monotonicity}
Let $\ca P(\real\hi d)$ be the set of probability measures on $\real\hi d$ and $\ca P\lo {ac}(\real\hi d)$ the set of probability measures absolutely continuous with respect to the Lebesgue measure. Let $\mu\in \ca P(\real\hi d)$ and $\nu\in \ca P(\real\hi d)$. A measurable map $T: \real\hi d\to \real\hi d$ is said to push $\mu$ to $\nu$ if for any measurable set $A\subseteq Y$, $\mu(T\inv(A))=\nu(A)$. This relation is frequently written as $\nu={T}_{\#}\mu$ or $\nu=\mu\circ T\inv$. Let $\ca T ( \mu, \nu)$ denote the class of all measurable functions $T$ such that $\nu = T \lo {\#} \mu$. Under the quadratic loss, Monge's optimal transport (MOT) seeks a member of $\ca T ( \mu, \nu)$ that reaches the infimum:
\begin{equation}\label{eq: monge}
    \inf \lo {T \in {\ca T (\mu, \nu)}} \left\{ \int \lo {\real \hi d} \| x - T(x) \| \hi 2 \mu (x) \right\}.
\end{equation}
If this infimum is achieved within $\ca T (\mu, \nu)$, then the minimizer is called an optimal transport map. However, the infimum may not be achievable within $\ca T (\mu, \nu)$ -- indeed,  $\ca T (\mu, \nu)$ can be an empty set in extreme cases. This limits the applicability of Monge's approach.

To address this limitation, \citet{kantorovich1948problem} introduced a relaxed version of Monge's problem by representing a transportation plan as a joint measure $\pi$ with marginals $\mu$ and $\nu$.  Let $\Pi(\mu, \nu)$ be the set of joint probability measures on $\real\hi d\times \real\hi d$ with marginals $\mu$ and $\nu$. The Kantorovich's problem seeks a $\pi$ in $\Pi(\mu, \nu)$ to minimize the total cost,  that is, 
\begin{equation}\label{eq: kp}
    W\lo2\hi2(\mu, \nu) = \inf\lo {\pi}\left\{\int\lo{\real\hi d\times\real\hi d} \|x - y\|\hi2 d\pi(x, y): \pi\in\Pi(\mu,\nu)\right\}.
\end{equation}
The square root of the minimum value of \eqref{eq: kp} is defined as the 2-Wasserstein distance, and a solution to \eqref{eq: kp} is called an optimal transport plan. The existence of the solution to \eqref{eq: kp} follows from \citet[Theorem 4.1]{villani2009optimal}.
Equivalently, we can express the square of the 2-Wasserstein distance as
\begin{align}\label{eq: semi_daul1}
    W\lo2\hi 2(\mu, \nu) = \int\|x\|\hi 2 d\mu(x) + \int\|y\|\hi2 d\nu -2\min \lo{\varphi\in \ca F} \left\{\int \varphi\, d\mu + \int \varphi\hi* d\nu\right\},
\end{align}
where $\ca F$ is the space of $L\lo1(\mu)$ convex function on $\real\hi d$, and $\varphi\hi *$ is the Legendre-Fenchel conjugate of $\varphi$, given by,
\begin{align}\label{eq: legendre_dual}
    \varphi\hi*(y) = \sup\lo{x\in \real\hi d}\{\la x, y\ra - \varphi(x)\},\,\, y \in \real\hi d.
\end{align}
Thus, solving \eqref{eq: kp} is equivalent to solving the semi-dual problem
\begin{align}\label{eq: semi_dual}
    \min\lo {\varphi\in \ca F}\int \left\{\varphi\, d\mu + \int \varphi\hi* d\nu\right\}.
\end{align}

The semi-dual problem \eqref{eq: semi_dual} connects with the Monge problem \eqref{eq: monge} through Brenier's Theorem \citep{brenier1991polar}, which establishes the existence, uniqueness, and intrinsic structure of the optimal transport map. \citet{mccann1995existence} extended this result to relax the second-order moment assumptions. 
\begin{proposition}[Brenier's Theorem]\label{prop: brenier}
Let $\mu$ and $\nu$ be two distributions on $\real\hi d$. 
\begin{itemize}
    \item[(1)] If $\mu$ is absolutely continuous with respect to the Lebesgue measure on $\real\hi d$, with support contained in a convex set $U$, then there exists a convex function $\varphi: U \to \real \cup \{+\infty\}$ such that $(\nabla \varphi)\lo{\#}\mu = \nu$. The function $\nabla\varphi$ is unique $\mu$-almost everywhere.
    \item[(2)] If, in addition, $\nu$ is absolutely continuous on $\real\hi d$ with support contained in a convex set $V$, then there exists a convex function $\varphi\hi*: V \to \real\cup\{+\infty\}$ such that $(\nabla\varphi\hi*)\lo{\#}\nu = \mu$. The function $\nabla\varphi\hi*$ is unique $\nu$-almost everywhere and equal to $(\nabla\varphi)\inv$ $\mu$-almost everywhere. That means, for almost every $x, y\in U\times V$,
    \begin{align*}
        (\nabla\varphi\hi*\circ\nabla\varphi)(x) = x, \quad(\nabla\varphi\circ\nabla\varphi\hi *)(y) = y.
    \end{align*}
\end{itemize}
\end{proposition}
Proposition \ref{prop: brenier} implies that a unique transport map with the form $\nabla\varphi$ exists between any absolutely continuous distributions. We refer to such a convex function $\varphi$ as the Brenier potential. 

On the real line $\real$, the gradients of convex functions are non-decreasing functions. When $d\ge 2$, \citet{rockafellar1966characterization} showed that the set of gradients of convex functions coincides with the set of cyclically monotonic functions defined below, which can be treated as a generalization of monotonicity to functions with more than one variables. 
\begin{definition}
Let $U$ be a nonempty subset of $\real\hi d$. A function $f$ is called cyclically monotone if, for every set of points $x_{1},\dots ,x_{n+1}\in U$ with $x_{n+1}=x_{1}$, it holds that $$ \sum _{k=1}^{n}\langle x_{k+1},f(x_{k+1})-f(x_{k})\rangle \geq 0.$$ Equivalently,
$$\sum_{k=1}^n\langle x_k, f(x_k)\rangle \geq \sum_{k=1}^n\langle x_k, f(x_{\sigma(k)})\rangle$$ for any permutation $\sigma$ of $\{1, 2, \dots, n\}$.
\end{definition}
As a special case, the linear function $f(x) = Ax$, where $A\in \real\hi{d\times d}$, is cyclically monotone if $A$ is symmetric and positive definite. When $d = 1$,  the definition above is equivalent to the usual notion of monotonicity.

\section{Cyclically Monotone Copula Gaussian Graphical\\ Model}\label{sec: CMC-GGM}
The structure of the multi-attribute graphical model leads us to propose a flexible semiparametric model for non-Gaussian data. As we have observed, non-Gaussianity may occur in node vectors of $X$ instead of occurring in the coordinates of $X$. Correspondingly, we let the transformations act jointly on the node vectors instead of coordinate-wise as in the classical copula transformation.

We define the Cyclically Monotone Copula Gaussian (CMCG) distribution as follows. Let  $X=(X\lo 1\trans, X\lo 2\trans,\dots, X\lo p\trans)\trans$ where, for $j=1,\dots, p$, $X\lo j =(X\lo{j1},\dots, X\lo{jd})$ is a random vector in $\real\hi d$.
\begin{definition}\label{def: CMCG_family}
We say $X$ follows a cyclically monotone copula Gaussian (CMCG) distribution if there exist cyclically monotone functions $\{T\lo j: \real\hi d\to \real\hi d\}\lo{j=1}\hi p$, such that $(T\lo 1(X\lo 1)\trans, \dots, T\lo p(X\lo p)\trans)\trans\sim N(0,\Sigma)$, where $\Sigma$ has structure \eqref{eq: cov_pre} with $\Sigma\lo{jj}= I\lo d$. 
\end{definition}
Let $T = (T\lo 1, \dots, T\lo p)$. We denote a random vector following CMCG distribution as $X \sim \mathrm{CMCG} (T,\Sigma)$. The CMCG family covers the copula Gaussian distribution family in \cite{liu2009nonparanormal} because the tensor product of univariate monotone functions is indeed a cyclically monotone function, as shown in the next proposition. In the following, for functions $g\lo1:\real\to\real, \dots, g\lo d: \real\to\real$, let $g\lo1\otimes\dots\otimes g\lo d$ denote the function from $\real\hi d\to\real\hi d$ such that $(g\otimes\dots\otimes g\lo d)(u\lo1, \dots, u\lo d) = (g\lo1(u\lo1), \dots, g\lo d(u\lo d))\trans$.
\begin{proposition}
If there exist monotone non-decreasing univariate functions $\{f\lo {js}: s=1,\dots d, j =1,\dots, p\}$ such that $(f\lo {11}(X\lo {11}), \dots, f\lo {pd}(X\lo{pd}))\sim N(0,\Sigma)$, then $X\sim \mathrm{CMCG}(T, \Sigma)$, where $T = (T\lo1, \dots, T\lo p)$ and $T\lo j = f\lo{j1}\otimes\dots\otimes f\lo{jd}$ for $j =1, \dots, p$.
\end{proposition}

For $j = 1,\dots,p$, let $P\lo {j}$ be the distribution of $X\lo j$. We assume $P\lo {j}$ to be absolutely continuous on $\real\hi {d}$. Then, by Brenier's theorem, there exists a unique cyclically monotone function $T\lo j$ such that $T\lo j(X\lo j)\sim N(0, I\lo d)$. In other words, the CMCG family allows the multi-dimensional marginal distributions to be any absolutely continuous distributions. As a comparison, the copula Gaussian model requires the multi-dimensional marginal distribution to be copula Gaussian. 

Let $X\sim\mathrm{CMCG} (T,\Sigma)$. Suppose the precision matrix $\Theta =\Sigma\inv$ has the same block structure as \eqref{eq: cov_pre}. We say $X$ follows a Cyclically Monotone Copula Gaussian Graphical Model (CMC-GGM) $\ca G(V, E)$ if \eqref{eq: graph_equi} holds. 

The joint cumulative distribution function of $X$ is given by:
\begin{align}\label{eq: cmc_cdf}
    F(x\lo{1},\dots,x\lo{p}) = \Phi\lo{\Sigma}(T\lo1(x\lo 1),\dots,T\lo p(x\lo p)),
\end{align}
where $\Phi\lo{\Sigma}$ is the multivariate normal cumulative distribution function with mean zero and covariance matrix $\Sigma$. If $\{T\lo j\}\lo{j=1}\hi p$ are differentiable, the joint probability density function of $X$ is given by:
\begin{align*}\label{eq:density}
    p\lo X(x)=\frac{1}{(2\pi)^{p/2}|\Sigma|^{1/2}}\exp\left\{-\frac{1}{2}T(x)\trans\Sigma\inv T(x)\right\}\prod_{j=1}^p\det(D\lo{T\lo j}(x\lo{j})),
\end{align*}
where $D\lo{T\lo j}$ is the Jacobian matrix of $T\lo j$ for $j=1, \dots, p$. This implies that the conditional independence $X_j\indep X_k|X_{-(j,k)}$ can still be characterized by $\Theta\lo{jk}=0$, which implies that relation \eqref{eq: graph_equi} is equivalent to 
\begin{align*}
    \Theta\lo {jk} = 0  \Leftrightarrow  (j, k)\notin E.
\end{align*}

In fact, any one-to-one transformations of the node vectors will preserve the original graph structure. However, a large class of transformations will make the model non-identifiable, prohibiting inference on the precision matrix. This is also noticed by \citet{bryan2021multirank} when designing semiparametric CCA models. The following proposition guarantees that the CMC-GGM is identifiable.
\begin{proposition}
If two random vectors $X\sim CMCG(T, \Sigma)$ and $\tilde X\sim CMCG(\tilde T, \tilde\Sigma)$ have the same distribution, then $\Sigma = \tilde \Sigma$, and $T = \tilde T$ almost everywhere. 
\end{proposition}

The CMC-GGM model extends the Copula-GGM model by assigning a multivariate normal score to each node vector. To construct rank-based estimators, we can solve optimal transport problems between the distribution of $X\lo j$ and the uniform distribution over the unit hypercube $[0,1]\hi{d}$. This is the same as the multivariate rank proposed in \citet{ghosal2019multivariate},
\citet{FAN2022}, and among others. The c.d.f of $X$ is given by
\begin{equation}\label{eq: ccmc}
    F(x\lo{1},\dots,x\lo{p}) = \Phi\lo{\Sigma}(\Phi\inv(R\lo1(x\lo 1))\trans,\dots,\Phi\inv(R\lo p(x\lo p)\trans),
\end{equation}
where $\Phi\lo{\Sigma}(\cdot)$ is the c.d.f of $N(0, \Sigma)$, $\Phi\inv(x\lo j) = (\Phi\inv(x\lo{j1}), \dots, \Phi\inv(x\lo{jd}))\trans$, and $R\lo j(\cdot)$ is the optimal transport map between $P\lo{j}$ and $U ([0,1]\hi{d})$.

For a Gaussian graphical model with a single attribute, $R\lo j(\cdot)$ is the common distribution function, and $\Phi\inv\circ R\lo j$ is a monotonic non-decreasing function. Therefore, model \eqref{eq: ccmc} and CMC-GGM \eqref{eq: cmc_cdf} are equivalent. However, this is not the case for the multi-attribute graphical model, as $\Phi\inv\circ R\lo j$ may not be a cyclically monotone function. In this paper, we focus on CMC-GGM because cyclically monotone functions offer the best transformations for preserving the relative information among data when transported to Gaussian. For further dicussions on Model \eqref{eq: ccmc}, please see the Supplementary Material.

\section{Estimation}\label{sec: estimation}
The plug-in procedure provides a direct approach to estimating the CMC-GGM. This approach was also used  \citet{liu2009nonparanormal} for the Copula-GGM and \cite{solea2020copula} for the copula functional graphical model. In this paper, we develop a two-step plug-in procedure to estimate CMC-GGM: in step 1, we estimate the transformation $T$ nonparametrically by solving  discrete-to-discrete optimal transport problems; in step 2, we use sparse estimation methods, including thresholding, group graphical lasso selection, and neighborhood vector-on-vector group lasso selection, to construct a sparse estimator of the blockwise precision matrix using transformed data.

\subsection{Estimation of the CMC Transformation}
Let $\ca X\lo n = \{X\hi {i}\}\lo{i=1}\hi n$ be i.i.d samples from $P\lo {X}$, where superscripts $i\in\{1, \dots, n\}$ is the sample index and $X\hi i=\{X\hi i\lo j: j =1, \dots, p\}$ where the subscript $j\in\{1, \dots, p\}$ is subvector index. We estimate the cyclically monotone transformation $T\lo j$ between $P\lo{j}$, the distribution of $X\hi i\lo j$, and the $d$-dimensional standard Gaussian distribution $Q$ by solving the following discrete-to-discrete OT problem. Let $\ca Z\lo n = \{Z\hi {1},\dots, Z\hi {n}\}$ be i.i.d samples drawn from $Q$. Define the empirical measures on $\ca {X}\lo n$ and $\ca {Z}\lo n$ as
\begin{align}\label{eq: phat_muhat}
    \hat P\lo j = \frac{1}{n}\sum_{i=1}^n\delta\lo {X\lo j\hi {i}},\quad \hat Q = \frac{1}{n}\sum\lo{i =1}\hi n\delta\lo{Z\hi {i}},
\end{align}
respectively. Solving the optimal transport problem between $\hat P\lo j$ and $\hat Q$ reduces to solving an assignment problem given by
\begin{equation}\label{eq: assign}
    \hat \sigma\lo j = \argmin\lo{\sigma\in A\lo n}\sum_{i = 1}^n\|X\lo j\hi {i}- Z\hi {\sigma(i)}\|\hi 2, 
\end{equation}
where $A\lo n$ is the set of all permutations of $\{1,2,\dots, n\}$. This is a combinatorial optimization problem and can be solved using the Hungarian algorithm (see, for example, \citet{jonker1988shortest}) with the worst computational complexity $\ca O(n\hi 3)$. The cyclically monotone transformation between $P\lo {j}$ and $Q$ is then estimated by $\hat T\lo j(X\lo j\hi i) = Z\hi{\sigma\lo j(i)}$ for $i = 1, \dots, n$.

To control the bias-variance trade-off in high-dimensional setting, we apply the Winsorization (or truncation) operator to the estimated transformation $\hat T\lo j$. Specifically, we define $\hat T\lo j\hii{w}(X\lo j) = (\hat T\lo {j1}\hii{w}(X\lo{j1}),\dots, \hat T\lo {jd}\hii{w}(X\lo{jd}))$, where 
\begin{equation}\label{eq: winsorization}
    \hat T\lo {js}\hii{w}(x) = \hat T\lo {js}(x)\bm 1\{|\hat T\lo {js}(x)| \le  \delta\lo n\} + \sign(\hat T\lo {js}(x))\delta\lo n\bm 1\{|\hat T\lo {js}(x)| \ge \delta\lo n\},
\end{equation}
is the 1-dimensional Winsorization operator with threshold $\delta\lo n$. In the classical copula graphical model, the winsorization operator is applied to the cumulative function $F(x)$, and the transformation is then estimated by $\Phi\inv(F\hii{w}(x))$. For example, \citet{klaassen1997efficient} consider using $\delta\lo n = n\inv$ and \citet{liu2009nonparanormal} suggest using $\delta\lo n = (4n\hi{1/4}\sqrt{\pi\log n})\inv$. In \eqref{eq: winsorization}, the winsorization operator is assigned to the transformed values of the estimated optimal transport map $\hat T$, which are samples from standard Gaussian. Therefore, we use the threshold $\delta\lo n \sqrt{2\log n}$, which provides comparable thresholding effects as the copula Gaussian setting. This choice is justified by \citet[Lemma 12.3]{abramovich2006adapting}, which states that $\Phi\inv(1-1/n) \le \sqrt{2\log n}$. This threshold enables us to derive the desired convergence rate in Section \ref{sec: asymptotic}.

We can also approximate the Gaussian distribution using the quasi-Monte Carlo methods (see \citet{deb2021multivariate} for more details). For example, we first take $\ca C\lo n = \{c\lo {1},\dots, c\lo {n}\}$ to be the $d$-dimensional Halton sequence of size $n$. The empirical distribution on $\ca C\lo n$ will be a discrete approximation of $U[0,1]\hi d$. Then the empirical distribution on $\{\Phi\inv(c\lo {i}), i = 1,\dots, n\}$, where $\Phi\inv$ is applied coordinatewise, can be considered as a discrete approximation of $N(0, I\lo d)$.  Since $\Phi\inv(\cdot)$ diverges very quickly when evaluated at a point close to $1$, we can also assign a Winsorization operator on the Halton sequence, resulting in the samples $\{Z\hi i = \Phi\inv((c\hii{w}\lo {i}))\}\lo{i=1}\hi n$. We do not find a significant difference between using quasi-Monte Carlo methods and the Monte Carlo method in simulation studies in Section \ref{sec: simu_cmc}.

\subsection{Sparse Estimation of the Precision Matrix}
In this subsection, we present methods for sparsely estimating the precision matrix using the estimated covariance matrix of the transformed data. Let $\hat \Sigma$ be the sample covariance matrices of the CMC-transformed data $\hat T\hii{w}(X)$ and $\hat \Theta$ be its inverse, with block structure
\begin{equation*}
\hat \Sigma = \begin{pmatrix}
    \hat \Sigma\lo{11}& \hat\Sigma\lo{12}&\dots&\hat\Sigma\lo{1p}\\
    \hat\Sigma\lo{21}& \hat\Sigma\lo{22}&\dots&\hat\Sigma\lo{2p}\\
    \vdots& \vdots & \ddots& \vdots\\
    \hat\Sigma\lo{p1} & \hat\Sigma\lo{p2}& \cdots & \hat\Sigma\lo{pp}
    \end{pmatrix}\,\, \text{and}\,\,
    \hat\Theta=\begin{pmatrix}
    \hat\Theta\lo{11}& \hat\Theta\lo{12}&\dots&\hat\Theta\lo{1p}\\
    \hat\Theta\lo{21}& \hat\Theta\lo{22}&\dots&\hat\Theta\lo{2p}\\
    \vdots& \vdots & \ddots& \vdots\\
    \hat\Theta\lo{p1} & \hat\Theta\lo{p2}& \cdots & \hat\Theta\lo{pp}
    \end{pmatrix}.
\end{equation*}
Here, the $(j,k)$-th block is
\begin{equation}\label{eq: hatsigma}
    \hat \Sigma\lo{jk} = {E\lo n[\hat T\hii{w}\lo j(X\lo j)\hat T\hii{w}\lo k(X\lo k)\trans]-E\lo n[\hat T\hii{w}\lo j(X\lo j)]E\lo n[\hat T\hii{w}\lo k(X\lo k)]},
\end{equation}
where $E\lo n[\cdot]$ is the empirical mean. We propose three ways to estimate the graph sparsely.

\paragraph{Thresholding.} We begin by computing the Tychonoff-regularized precision matrix as
\begin{align*}
  \hat \Theta \hii{r} = (\hat \Sigma + \eta I\lo {dp})\inv,
\end{align*}
where $\eta$ is a tuning parameter, and superscript $r$ indicates regularization. We then estimate the edge set $E$ by
\begin{align*}
    \hat E(\epsilon\lo n) = \{(j,k)\in V\times V: \|\hat \Theta\hii{r}\lo{jk}\|\lo{\mathrm{F}} >\epsilon\lo n\},
\end{align*}
where $\{\epsilon\lo n\}$ is a positive sequence with $\epsilon\lo n\downarrow 0$ and $\|\cdot\|\lo{\mathrm{F}}$ denotes the Frobenius norm. The thresholding estimator is direct and easy to implement but less accurate than penalized estimation methods in most settings.

\paragraph{Group glasso.} This approach minimizes the penalized negative Gaussian loglikelihood:
\begin{align}\label{eq: glasso}
    L\lo n: \real\hi{dp\times dp}\to \real, \,\,\Theta \to  -\log\det(\Theta) + \trace(\Theta\hat \Sigma)+\lambda\lo n\sum_{k\neq j}\|\Theta\lo{jk}\|\lo{\mathrm{F}},
\end{align}
where $\lambda\lo n$ is a tuning parameter. The precision matrix $\Theta$ is then estimated by minimizing $L\lo n(\cdot)$ over the set of all positive semidefinite $dp\times dp$ matrix. Several efficient algorithms have been developed to minimize the penalized negative Gaussian loglikelihood with group penalty, such as \cite{kolar2014graph} and \citet{tugnait2021sparse}, among others. We adopt the block coordinate descent algorithm in \citet{qiao2019functional} to solve \eqref{eq: glasso}. 

\paragraph{Neighborhood vector-on-vector group lasso selection.}
We perform vector-on-vector regression separately for each $j = 1,2,\dots, p$, using $\hat T\lo j(X\lo j)$ as the response and $p-1$ remaining subvectors in $\hat T(X)$ as predictors. Let $\varsigma$ be a mapping that reorganizes the indices such that $\{\varsigma(1), \dots, \varsigma(p-1)\} = \{1,\dots, p\} /\{j\}$. For simplicity, we write $Y =  \hat T\lo j(X\lo j) \in \real\hi{d}$ and $\tilde X = (\tilde X\lo 1,\dots, \tilde X\lo {p-1}) = (\hat T\lo {\varsigma(1)}(X\lo {\varsigma(1)}),\dots, \hat T\lo {\varsigma(p-1)}(X\lo {\varsigma(p-1)}))\in \real\hi{d\times (p-1)d}$. Let $B = (B\lo 1\trans, \dots, B\lo {p-1}\trans)\trans\in \real\hi{d(p-1)\times d}$, where $B\lo {k}\in \real\hi{d\times d}$ for $k = 1,\dots, p-1$. Then the vector-on-vector regression model can be written as:
\begin{equation}\label{eq: neighbor_lasso1}
    Y = \sum_{k=1}^{p-1}\tilde X\lo kB\lo k + \varepsilon = \tilde X B + \varepsilon,
\end{equation}
where $\varepsilon\in \real\hi {d}$ is an error vector with mean $0$ and is independent of $\tilde X$. After vectorization, \eqref{eq: neighbor_lasso1} can be written as 
\begin{equation}
    \vec(Y) = (I\lo d\otimes \tilde X)\cdot \vec(B) + \vec(\varepsilon),
\end{equation}
where $\vec(\cdot)$ vectorizes an $d \times d$ matrix by stacking the columns of the matrix into a $d\hi2\times 1$ vector and $\otimes$ is the Kronecker product. We then estimate regression parameters by minimizing the squared residuals with group lasso penalty, given by:
\begin{align}\label{eq: neighbor_lasso}
    \hat B = \underset{{B \in \real\hi{d(p-1)\times d}}}{\argmin} \left\{\frac{1}{2n}\sum\lo{i=1}\hi n\left\|\vec(Y\hi i) - (I\lo{d}\otimes \tilde X\hi i)\vec(B)\right\|\hi 2+\lambda\lo n\sum_{j=1}\hi {p-1}\|\vec(B\lo k)\|\lo{2}\right\},
\end{align}
where $\lambda\lo n$ is a tuning parameter. We estimate the support set for node $j$ as:
\begin{align*}
    \hat {\ca N}\lo j = \{\sigma(k): k = 1,\dots, p-1, \|\hat B\lo k\|> 0\}.
\end{align*}
After estimating the neighborhood for each node, we construct an estimated edge set $\hat E$ by aggregating $\{\hat {\ca N}\lo j\}\lo{j=1}\hi p$ via intersection or union. We use the Groupwise Majorization Descent (GMD) algorithm in \citet{yang2015fast} to solve \eqref{eq: neighbor_lasso}. 

\paragraph{Selection of tuning parameters.}
All methods above require choosing tuning parameters to control the sparsity of the estimated graph.  The sparsity level can be controlled by $\epsilon\lo n$ for thresholding, and by $\lambda\lo n$ for glasso and neighborhood selection. Common approaches for tuning parameter selection include the Akaike information criterion (AIC), Bayesian information criterion (BIC), cross-validation, and stability selection \citep{meinshausen2010stability}. In the synthetic data experiments, to ensure a fairer comparison among different methods, we fit each method over a range of tuning parameters and generate ROC curves. We then compute the associated area-under-curve (AUC) values. For the data applications, we suggest using the BIC to select tuning parameters for the group glasso method, which takes the following form:
\begin{align}\label{eq: bic}
    \mathrm{BIC}(\lambda\lo n) = \trace(\hat \Sigma\hat\Theta) - \log|\hat \Theta| + \frac{\log(n)}{n}m\hi2\left(\frac{1}{2}\sum\lo{j\neq k}1\{\hat\Theta\lo{jk}\neq 0\} + p\right).
\end{align}
However, when the sample size is small, BIC may not lead to a reasonable graph. To address this issue, we suggest a method similar to the stability selection \citep{meinshausen2010stability}. First, we fit a relatively dense graph with the sparsity chosen by cross-validation or domain knowledge. Next, we refit the model using bootstrap samples 50 times and select the stable edges that appeared in at least $90\%$ of the replications.  


\section{Consistency and Convergence Rate}\label{sec: asymptotic}
In this section, we establish the consistency and convergence rate for the estimator of the CMC-GGM. Recall that each $X\lo j$ in $X$ is distributed as $P\lo j$, which is dominated by the Lebesgue measure in $\real\hi d$. Let $Q = N(0, I\lo d)$. By Brenier's theorem, for each $j=1,\dots, p$, there exists a convex potential function $\varphi\lo j$ such that the optimal transport map $T\lo j$ between $P\lo j$ and $Q$ can be written as $T\lo j = \nabla\varphi\lo j$. Let $\varphi\lo j\hi*$ be the Legendre-Fenchel dual of $\varphi\lo j$, as defined in \eqref{eq: legendre_dual}. Then, $\nabla \varphi\lo j\hi*$ is the optimal transport map transporting $Q$ to $P\lo j$. Let $\hat P\lo j$ and $\hat Q$ be the empirical measures defined in \eqref{eq: phat_muhat}. Let $\hat T\lo j$ be the estimates obtained from \eqref{eq: assign}.


Several recent works, such as \citet{deb2021multivariate} and \citet{hallin2021distribution}, focus on establishing the convergence rate of the discrete-to-discrete estimator of the optimal transport map. In our model, both the source and target distributions can have unbounded supports, where the results in \citet{manole2021plugin} and \citet{deb2021multivariate} cannot be directly applied. Similar to \citet{manole2021plugin}, we introduce the following regularity condition on the Brenier potential functions $\{\varphi\lo j\}\lo {j=1}\hi p$ to give a stability bound of the estimated transformations. For $j = 1, \dots, p$, we assume
\begin{itemize}
    \item[(A1)] $\nabla \varphi\lo j$ is Lipschitz continuous with Lipschitz constant $\rho$;
    \item[(A2)] $\varphi\lo j$ is strongly convex with parameter $1/\rho$.
\end{itemize}
Assumptions (A1) and (A2) guarantee that for any two points $x, y\in \real\hi d$,
\begin{equation*}
    \frac{1}{\rho}\|x-y\|\hi 2\le \langle \nabla \varphi\lo j(x) - \nabla\varphi\lo j(y), x-y\rangle \le {\rho}\|x-y\|\hi 2.
\end{equation*}
\begin{remark}
By \citet[Theorem 4.2.1, 4.2.2]{hiriart2004fundamentals}, the Legendre-Fenchel
dual $\varphi\lo j\hi*(\cdot)$ is strongly convex with parameter $1/\rho$ if and only if $\nabla\varphi\lo j(\cdot)$ is $\rho$-Lipschitz continuous. Therefore, assumption (A2) is equivalent to $\nabla\varphi\lo j\hi*$ being $\rho$-Lipschitz.
\end{remark}
\begin{remark}
We note that assumption (A1) implies (A2) when both source and target distributions are defined in a compact set  and have densities that are bounded above and below. However, when the support of $P\lo j$ and $Q$ are unbounded, assumption (A2) becomes more restrictive. Nevertheless, under certain regularity conditions on the source distribution $P\lo j$, we can still guarantee (A2). For example, according to Caffarelli contraction theorem \citep{caffarelli2000monotonicity}, if $P\lo j$ is a uniformly log-concave measure of the form $e\hi{-V}dx$, where $V$ has Hessian matrix $\ge \alpha I\lo d$, then $\nabla\varphi\lo j\hi*$ is $\alpha\hi{-1/2}$-Lipschitz and hence $\varphi\lo j$ is $\alpha\hi{-1/2}$-strongly convex. For further extensions, see \citet{colombo2021bounds} and \citet{manole2021sharp}.
\end{remark}
We first establish the following stability bound. A similar result for the semi-discrete OT setting is derived in \citet[Theorem 6]{manole2021plugin} and \citet[Theorem 2.1]{deb2021rates}.
\begin{lemma}\label{lemma: stability bound}
Suppose assumption (A1) holds with constant $\rho > 0$, then, for $j = 1, \dots, p$, we have 
\begin{equation}\label{eq: stability}
    \frac{1}{n}\sum_{i = 1}^n \|\hat T\lo j(X\hi i\lo j) - T\lo j(X\hi i\lo j)\|\hi 2  \le \rho \left\{W\lo 2\hi 2(\hat P\lo j, \hat Q) - W\lo2\hi2(\hat P\lo j, \bar Q\lo j) - \int g\lo j\, d(\hat Q - \bar Q\lo j)\right\}, 
\end{equation}
where $g\lo j(\cdot) = \|\cdot\|\hi2 - 2\varphi\lo j\hi*(\cdot)$ and $\bar Q\lo j = (\nabla\varphi\lo j)\lo \#(\hat P\lo j)$. If, in addition, assumption (A2) holds, then 
\begin{align}\label{eq: stability_2}
     \frac{1}{\rho\hi2} W\hi2\lo2(\hat Q, \bar Q\lo j)\le \frac{1}{n}\sum_{i = 1}^n \|\hat T\lo j(X\hi i\lo j) - T\lo j(X\hi i\lo j)\|\hi 2  \le \rho\hi2 W\hi2\lo2(\hat Q, \bar Q\lo j).
\end{align}
\end{lemma}
Based on the stability bounds in Lemma \ref{lemma: stability bound}, we establish a $0$-concentration inequality for  $\|\hat T\lo j - T\lo j\|\lo{L\lo 1(\hat P\lo j )}$ in Lemma \ref{lemma: concentration_t_hat_t}. Since we use Monte Carlo methods to generate discrete samples from $Q$, both the randomness of $X$ and $Z$ are considered in the concentration inequality. Throughout the following, we use $C$ to denote general constants independent of $n$ and $p$ but may change from one place to another. Let $\zeta\lo d = 1/2$ if $d = 4$ and $0$ otherwise.

\begin{lemma}\label{lemma: concentration_t_hat_t}
Under assumptions (A1) and (A2),  for $j = 1, \dots, p$, we have 
\begin{align*}
    \mathbb P\left(\frac{1}{n}\sum_{i = 1}^n \left\|\hat T\lo j(X\hi i\lo j)-T\lo j (X\hi i\lo j)\right\|\ge \varepsilon\right)\le  \exp\left\{-Cn\left[(\varepsilon - Cn\hi{-\frac{1}{4\lor d}}(\log n)\hi{\zeta\lo{d}})\lo+\right]\hi 2\right\},
\end{align*}
where $a\lo+ = aI\{a > 0\}$.
\end{lemma}
Let $T\hii{w}\lo j(x)$ and $\hat T\hii{w}\lo j(x)$ be the winsorized versions of $T\lo j(x)$ and $\hat T\hii{w}(x)$, respectively, with a threshold of $\sqrt{2\log n}$. The concentration inequality in Lemma \ref{lemma: concentration_t_hat_t} applies to  $\hat T\hii{w}$ and $T\hii{w}$ by the observation that $\|\hat T\lo j(X\hi i\lo j)-T (X\hi i\lo j)\|\ge \|\hat T\hii{w}\lo j(X\hi i\lo j)-T\hii{w} (X\hi i\lo j)\|$ for any $1\le i\le n$.
\begin{coro}\label{coro: concentration_win}
Under assumptions (A1) and (A2), for $j = 1, \dots, p$, we have
\begin{align*}
    \mathbb P\left(\frac{1}{n}\sum_{i = 1}^n \left\|\hat T\hii{w}\lo j(X\hi i\lo j)-T\hii{w} (X\hi i\lo j)\right\|\ge \varepsilon\right)\le  \exp\left\{-Cn\left[(\varepsilon - Cn\hi{-\frac{1}{4\lor d}}(\log n)\hi{\zeta\lo{d}})\lo+\right]\hi 2\right\}.
\end{align*}
\end{coro}

We now show that the estimator $\hat \Sigma$ converges to $\Sigma$ blockwise when $p$ grows at an exponential rate of $n$. Recall that $\hat\Sigma$ is an estimator after the winsorization.
\begin{theorem}\label{thm: main}
For any $\varepsilon > 0$, there exists a constant $C>0$ such that  
\begin{align*}
     \mathbb P\left(\left\|\hat \Sigma\lo{jk} - \Sigma\lo{jk}\right\|\lo {\mathrm F}\ge \varepsilon\right)\le C\exp\left\{-C\frac{n}{d\log\hi2 n}\left[(\varepsilon - Cn\hi{-\frac{1}{4\lor d}}(\log n)\hi{\zeta\lo{d} + \frac{1}{2}})\lo+\right]\hi2\right\},
\end{align*}
and consequently,
\begin{align*}
     \mathbb P\left(\max\lo{1\le j, k\le p}\left\|\hat \Sigma\lo{jk} - \Sigma\lo{jk}\right\|\lo {\mathrm F}\ge \varepsilon\right)\le Cp\hi 2\exp\left\{-C\frac{n}{d\log\hi2 n}\left[(\varepsilon - Cn\hi{-\frac{1}{4\lor d}}(\log n)\hi{\zeta\lo{d} + \frac{1}{2}})\lo+\right]\hi2\right\}.
\end{align*}
\end{theorem}
The proof of Theorem \ref{thm: main} can be found in the Supplemental Material. As can be seen from the proof, the tail bound of the mean concentration of $\|\hat \Sigma\lo{jk} - \Sigma\lo{jk}\|\lo {\mathrm F}$ has an order of ${\log n}/\sqrt{n}$, that is, $(\|\hat \Sigma\lo{jk} - \Sigma\lo{jk}\|\lo {\mathrm F} - \mathbb E[\|\hat \Sigma\lo{jk} - \Sigma\lo{jk}\|\lo {\mathrm F}]) = O\lo p({\log n}/\sqrt{n})$. However, this rate is dominated by the convergence rate of the mean $\mathbb E[\|\hat \Sigma\lo{jk} - \Sigma\lo{jk}\|\lo {\mathrm F}]$, which has an order of $(\log n)\hi{\zeta\lo d + \frac{1}{2}}/(n\hi{\frac{1}{4\lor d}})$. We compare Theorem \ref{thm: main} with the convergence rate of the Copula-GGM. For example, see \citet[Theorem 3]{mai2022coordinatewise} for the convergence rate of the normal score estimator and winsorization estimator of the Copula-GGM. We see that the Winsorization estimator $\hat \Sigma$ of CMC-GGM has the same rate of tail bound as the Copula-GGM, but a slower rate of mean convergence, which is also of order $\log n/\sqrt{n}$ for Copula-GGM. This slower rate of mean convergence is due to the curse of dimensionality issue that arises when solving $d$-dimensional OT problems. Furthermore, when $\log p = o(n/\log\hi2 n)$, we have $\max\lo{1\le j, k\le p}\|\hat \Sigma\lo{jk} - \Sigma\lo{jk}\|\lo {\mathrm F} = O\lo p((\log n)\hi{\zeta\lo d + \frac{1}{2}}/(n\hi{\frac{1}{4\lor d}})) = o\lo p(1)$. This indicates that the dimension limit can be at $\log p = O(n\hi\tau)$ with $0<\tau<1$.

To further establish the consistency of the group glasso estimator \eqref{eq: glasso} for graph estimation, we introduce a group version of the irrepresentable condition, which is also used in \citet{kolar2014graph} and \citet{qiao2019functional}, as an extension of the irrepresentable condition in \citet{ravikumar2011high}. Let $A = (A\lo {jk})\lo{j,k=1}\hi p$ be an $\real\hi{pd\times pd}$ block matrix, where $A\lo{jk}\in \real\hi {d\times d}$ for $j,k=1,\dots, p$. We define the blockwise norms $\|A\|\hii{d}\lo{\infty} = \max\lo{1\le j\le p}\sum\lo{k=1}\hi p\|A\lo{jk}\|\lo{\mathrm{F}}$, $\|A\|\hii{d}\lo{\max} = \max\lo{1\le j,k \le p}\|A\lo{jk}\|\lo{\mathrm{F}}$, regarding them as the block versions of matrix $\ell\lo\infty$-norm and maximum norm, respectively. The superscript $(d)$ indicates the length of the block. Let $\tilde E = E\cup \{(1,1), \dots, (p, p)\}$ be the augmented edge set and $\tilde E\hi c$ be its complement. Let $H = \Theta\inv\otimes \Theta\inv\in \real\hi{(pd)\hi2\times(pd)\hi2}$, where $\otimes$ is the Kronecker product. The matrix $H$ is the Hessian operator of the loglikelihood function evaluated at the true $\Theta$. For index set $J, J\hi'\subseteq\{1, \dots, p\}$, let $H\lo{JJ\hi'}\in \real\hi{d\hi2|J|\times d\hi2|J\hi'|}$ be the submatrix of $H$ with row and column blocks in $J$ and $J\hi'$. We assume that the following irrepresentable-type condition holds.
\begin{itemize}
    \item[(A3)] There exists a constant $0\le \alpha<1$ such that
    \begin{align*}
        \|H\lo{\tilde E\hi c \tilde E}(H\lo{\tilde E\tilde E})\inv\|\lo{\infty}\hii{d\hi2} \le 1-\alpha.
    \end{align*}
\end{itemize}
Let $\kappa\lo{\Sigma} = \|\Sigma\|\lo{\infty}\hii{d}$ and $\kappa\lo H = \|H\inv\|\lo{\infty}\hii{d\hi2}$. Let $s$ be the maximal degree of nodes
in $\ca G$. Define the following quantities:
\begin{align*}
    \delta\lo 1(n) = \frac{s\sqrt{d\log p}\log n}{\sqrt{n}}, \,\,\text{and}\,\, \delta\lo2(n) = \frac{(\log n)\hi{\zeta\lo d+ \frac{1}{2}}}{n\hi{\frac{1}{4\lor d}}}.
\end{align*} 
Under conditions (A1), (A2), and (A3), we have the selection consistency of the CMC-GGM in the following theorem. 
\begin{theorem}\label{thm: selection}
Let $\lambda\lo n \asymp \delta\lo1(n)\land \delta\lo2(n)$. Then,  as $\delta\lo1(n)\land \delta\lo2(n)\to 0$, $\|\hat \Theta - \Theta\|\lo{\max}\hii{d} = o\lo p(1)$. Consequently, the estimated graph $\hat G$ agrees with the true graph $G$ with high probability, that is, $\mathbb P(\hat G = G) \to 1$.
\end{theorem}

In Theorem \ref{thm: selection}, we see that $\delta\lo1(n)$ corresponds to the tail bound while $\delta\lo2(n)$ corresponds to the mean error bound. When $\log p \prec n\hi{(1-\frac{2}{4\lor d})}(\log n)\hi{(2\zeta\lo d - 1)}$, $\delta\lo 1(n)\prec \delta\lo 2(n)$, indicating that the mean error bound dominates the tail error bound. When $\log p \succ n\hi{(1-\frac{2}{4\lor d})}(\log n)\hi{(2\zeta\lo d - 1)}$, $\delta\lo 1(n)\succ \delta\lo2(n)$, indicating that the tail error bound dominates the mean error bound, and the convergence rate is the same as copula Gaussian model in \citet[Theorem 4]{mai2022coordinatewise}. The theorem also indicates that the dimension limit can be at most $\log p = O(n/\log\hi2 n)$.

The selection consistency using CMC transformed score can also be established with neighborhood group lasso selector, following the proof in, for example, \citet{zhao2021high}. We skip this part due to space limitation.

\section{Projected Cyclically Monotone Copula for High Dimensional Attributes}
\label{sec: pcmc}
\subsection{Projected Cyclically Monotone Copula}
In Section 3.1, we show that estimating the CM transformations $\{T\lo j\}\lo{j=1}\hi p$ involves solving discrete-to-discrete OT problems. However, this estimation method suffers from curse of dimensionality. In fact, the minimax convergence rate of the estimated OT map between $d$-dimensional discrete measures is $O(n\hi{-1/d})$ \citep{niles2022estimation,fournier2015rate}. Theorems \ref{thm: main} and \ref{thm: selection} also indicate that the convergence rate of graph estimation is limited by the dimension of vectors on each node.
However, when the dimension of the vector is high, it is reasonable to assume the non-Gaussianity only appears on a $r$-dimensional subspace, where $r< d$. Once the $r$-dimensional subspace is properly estimated, the convergence rate can be improved from $O(n\hi{-1/d})$ to $O(n\hi{-1/r})$. To achieve this, we adopt the idea in projection robust OT method \citep{paty2019subspace} to develop a projected cyclically monotone copula model that improves the estimation accuracy.  

For $j = 1, \dots, p$, we assume that there exists an orthogonal matrix $\Gamma\lo j = (U\lo j, V\lo j)\in \real \hi{d\times d}$, where $U\lo j\in\real\hi{d\times r}$ and $V\lo j\in \real\hi{d\times (d-r)}$, such that $P\lo j$ and $Q$ only differ on an $r$-dimensional subspace spanned by $U\lo j$. Let $Z\sim N(0, I\lo d)$. By Brenier's Theorem, the optimal transport between $\Gamma\trans X\lo j$ and $\Gamma\trans Z$ can be written as $(T\lo j, \mathrm{id}\lo{\real\hi{d-r}})$, where $T\lo j: \real\hi r\to\real\hi r$ is the optimal transport from $U\trans X\lo j$ to $U\trans Z$. Thus, the induced transformation $S\lo j$ from $X\lo j$ to $Z$ can be written as 
\begin{align*}
    S\lo j: =  \Gamma\circ(T\lo j, \mathrm{id}\lo{\real\hi{d-r}})\circ \Gamma\trans = U\lo j\circ T\lo j\circ U\lo j\trans + V\lo jV\lo j\trans.
\end{align*}
We now define the Projected Cyclically Monotone Copula Gaussian (PCMCG) family.
\begin{definition}\label{def: pCMCG_family}
A random vector $X=(X\lo 1\trans, X\lo 2\trans,\dots, X\lo p\trans)\trans\in \real\hi {dp}$ with subvector $X\lo j \in\real \hi d$, follows a projected cyclically monotone copula Gaussian (PCMCG) distribution if there exist cyclically monotone functions $T = \{T\lo j: \real\hi r\to \real\hi r\}\lo{j=1}\hi p$ and orthogonal matrices $\Gamma = \{\Gamma\lo j\in \real\hi{d\times d}: \Gamma\lo j = (U\lo j, V\lo j), U\lo j\in \real\hi {d\times r}, V\lo j\in \real\hi{d\times(d-r)}, \Gamma\lo j\trans \Gamma\lo j = \Gamma\lo j \Gamma\lo j\trans = I\lo d\}\lo {j=1}\hi p$ such that $(S\lo 1(X\lo 1)\trans, \dots, S\lo p(X\lo p)\trans)\trans\sim N(0,\Sigma)$, where $S\lo j = U\lo j\circ T\lo j\circ U\lo j\trans + V\lo j V\lo j\trans$ and $\Sigma$ has structure \eqref{eq: cov_pre} with $\Sigma\lo{jj}= I\lo d$. 
\end{definition}

Let $Z = S\lo j(X\lo j)$. By the fact that $Z\sim N(0, I\lo d)$, we know that $U\lo j\trans Z$ and  $V\lo j\trans Z$ are independent, implying that $U\lo j\trans X\lo j$ and $V\lo j\trans X\lo j$ are independent. Since $S\lo j, j=1,\dots, p$ are cyclically monotone functions, they are unique optimal transport transformations from $P\lo j$ to $Q$ almost surely. Hence, the PCMCG family can be treated as a special case of the CMCG family in Definition \ref{def: CMCG_family}.  

Let $\ca S\lo j$ be the space spanned by $U\lo j$ and $\ca S\lo j\oc$ be its orthogonal complement spanned by $V\lo j$. Denote by $P\lo {\ca S\lo j} = U\lo jU\lo j\trans$ the projection matrix onto the space spanned by $U\lo j$.  We note that for a different choice of bases $U\lo j$, the function $T\lo j$ may change, but the composition map $U\lo j\circ T\lo j\circ U\lo j\trans$ will be invariant. This is guaranteed because $U\lo j\circ T\lo j\circ U\lo j\trans$ is the optimal transport from the distribution of $P\lo{S\lo j}X\lo j$ to the distribution of $P\lo{S\lo j} Z$. Similarly, $S\lo j$ is invariant because $S\lo j$ is the optimal transports from $P\lo j$ to $Q$. By Brenier theorem, they are unique almost surely.

Define an equivalence relation $T\sim \tilde T$, if and only if there exists an orthogonal matrix $A\in \real\hi{r\times r}$ such that $T = A\circ \tilde T\circ A\trans$. Let $[T]$ be the equivalence class of $T$ over the set of cyclically monotone functions. We define the map $P\lo {\ca S}([T]) = A\circ T\circ A\trans$ as the projected cyclically montone transformation of $[T]$ onto subspace $\ca S$, where $A$ is any basis matrix spanning $\ca S$. $P\lo {\ca S}([T])$ is well-defined because it is invariant for the choice of basis $A$ and representative element $[T]$. Therefore, the transportation map defined in Definition \ref{def: pCMCG_family} can be written as 
\begin{equation}\label{eq: trans_pCMCG}
    S\lo j = P\lo{\ca S\lo j}([T\lo j]) + P\lo{\ca S\lo j\hi\oc},
\end{equation} 
where $\ca S\lo j$ is the space spanned by $U\lo j$ and $P\lo{\ca S\lo j\hi\oc}$ is the usual projection matrix on subspace $\ca S\lo j\hi\oc$. Let $S = (S\lo1, \dots, S\lo p)$ and $\ca S = (\ca S\lo1, \dots, \ca S\lo p)$. Therefore, for a random vector $X$ defined by Definition \eqref{def: pCMCG_family}, we write $X\sim \mathrm{PCMCG}(\ca S, S, \Sigma)$. 

Similarly, for a radnom vector $X\sim \mathrm{PCMCG}(\ca S, S, \Sigma)$, we say it follows a CCMC-GGM with graph $\ca G=(V, E)$ when $\Theta\lo {jk} = 0$ is equivalent to $ (j, k)\notin E$. The following proposition guarantees that the PCMC-GGM is identifiable. 
\begin{proposition}
If two random vectors $X\sim \mathrm{PCMCG}(\ca S, S, \Sigma)$ and $\tilde X\sim \mathrm{PCMCG}(\tilde {\ca S}, \tilde S, \tilde\Sigma)$ have the same distribution, then $\ca S = \tilde{\ca S}$, $\Sigma = \tilde \Sigma$, and $S =\tilde {S}$ almost everywhere. 
\end{proposition}

\subsection{Estimation}
Define $U\hi*: \real\hi d\to \real\hi k$ to be the linear transformation associated with $U$ by $U\hi*(x) = U\trans x$. To estimate the non-Gaussianity subspace $\ca S\lo j$, we consider the worst possible optimal transport cost over all possible $r$-dimensional subspace, that is,
    \begin{equation*}
        U\lo j = \argmax\lo{U\in \mathrm{St}(d, r)} W\lo2(U\hi*\lo{\#}P\lo j, U\hi*\lo{\#}Q),
    \end{equation*}
where $\mathrm{St}(d, r) := \{U\in \real\hi{d\times r}: U\trans U = I\lo r\}$ denotes the Stiefel manifold. At the sample level, we solve the max-min optimization problem
\begin{equation}\label{eq: max-min}
   \max\lo{U\in \mathrm{St}(d, r)}\min\lo{\pi\in\Pi}\sum\lo{s=1}\hi n\sum\lo{t=1}\hi n\pi\lo{s, t}\|U\trans X\lo j\hi s- U\trans Z\hi t\|\hi 2,
\end{equation}
where $\Pi:= \{\pi\in \real\hi{n\times n}\lo +: \pi1\lo n = 1\lo n, \pi\trans 1\lo n= 1\lo n\}$ denotes the transportation polytope. Here, we take the Kontorovich formulation of the OT problem for computing efficiency. We denote the solution of \eqref{eq: max-min} as $\hat U\lo j$ and $\hat \pi\lo j$. The optimal transport plan $\pi$ indeed defines an optimal transport map in the sense that 
\begin{align*}
    \pi\lo{st} = I\{t = \sigma\lo j(s)\} /  n, \quad \text{for all} \,\, 1\le s, t\le n,
\end{align*}
where $\sigma\lo j(\cdot)$ is a permutation of $\{1, \dots, n\}$. Define $\hat T\lo j (\hat U\lo jX\hi i\lo j) = \hat U\lo jZ\hi{\sigma\lo j(i)}, i = 1, \dots, n$. The final transformations are then estimated by: 
\begin{equation}\label{eq: hat_S}
\hat S\lo j(X\lo j\hi i) = \hat U\lo j\hat T\lo j(\hat U\lo j\trans X\hi i\lo j) + \hat V\lo j\hat V\lo j\trans X\hi i\lo j, \quad \text{for}\,\, i = 1,\dots, n,\,\, j=1,\dots, p.    
\end{equation}

Define $\hat P\lo j$ and $\hat Q$ same as in \eqref{eq: phat_muhat}. Let $P\lo {j,\hat U\lo j}$ and $Q\lo {\hat U\lo j}$ be the distribution of $\hat U\lo j\trans X\lo j$ and $\hat U\lo j\trans Z$, respectively. Similarly, define the empirical measures as:
\begin{align*}
    \hat P\lo{j, \hat U\lo j} = \frac{1}{n}\sum_{i=1}^n\delta\lo {\hat U\lo j\trans  X\lo j\hi {i}},\quad \hat Q\lo{\hat U\lo j} = \frac{1}{n}\sum\lo{i =1}\hi n\delta\lo{\hat U\lo j\trans Z\hi {i}}.
\end{align*}
Then $\hat T\lo j$ is the optimal transport from $\hat P\lo{j, \hat U\lo j}$ to $\hat Q\lo{\hat U\lo j}$. We note that $\hat S$ defined in \eqref{eq: hat_S} does not transform $\hat P\lo j$ to $\hat Q$. Instead, $\hat S$ is the optimal transport from $\hat P$ to $\hat Q\hi*$, where
\begin{align*}
    \hat Q\hi*= \frac{1}{n}\sum_{i=1}^n\delta\lo {\hat U\lo j \hat U\lo j\trans Z\hi{\sigma\lo j(i)} + \hat V\lo j\hat V\lo j\trans X\hi i}.
\end{align*}

In practice, we first estimate the projection matrix $U\lo j$ by solving \eqref{eq: max-min} with entropy penalty to speed computation, that is,
\begin{equation}\label{eq: max-min_sinkhorn}
   \max\lo{U\in \mathrm{St}(d, k)}\min\lo{\pi\in\Pi}\sum\lo{s=1}\hi n\sum\lo{t=1}\hi n\pi\lo{s, t}\|U\trans X\lo j\hi s- U\trans Z\hi t\|\hi 2 - \eta H(\pi),
\end{equation}
where $\eta > 0$ is a tuning parameter and $H(\pi) := -\la \pi, \log(\pi) - 1\lo n1\lo n\trans\ra$. To solve \eqref{eq: max-min_sinkhorn}, we adopt the Riemannian block coordinate descent (RBCD) algorithm proposed in \citet{huang2021riemannian}. To avoid the bias of estimating $\pi$ due to the entropy penalty, we only obtain $\hat U\lo j, j =1,  \dots, p$ from \eqref{eq: max-min_sinkhorn} and then solve an exact OT problem using the projected data in a follow-up step.

\subsection{Convergence Rate}
We next develop the convergence rate of the PCMCG model for $r = 1$, where there exists one principal non-Gaussian direction. For $j = 1,\dots, p$, we assume that $P\lo {j, U\lo j}$ is absolutely continuous to the Lebesgue measure, and thus, there exists a convex potential function $\psi\lo j$, such that $T\lo j = \nabla\psi\lo j$ is the optimal transport from $P\lo {j, U\lo j}$ to $Q\lo {U\lo j}$. Let $\psi\lo j\hi*$ be the Legendre-Fenchel dual of $\psi\lo j$, defined in \eqref{eq: legendre_dual}. To ensure the consistency of estimating the projected subspace $U_j$, we introduce two additional regularity conditions. First, we introduce the log Sobolev inequality to characterize the tail behavior of $X$ as follows: 
\begin{definition}
    A probability measure $\mu$ on $\real\hi d$ is said to satisfy a log Sobolev inequality with constant $\kappa\hi2$ if 
    \begin{align}\label{eq: log_sob}
        \int f\hi2\log f\hi2 \,d\mu - \int f\hi2\, d\mu\log \left(\int f\hi2\, d\mu\right)\le 2\kappa\hi2\int \|\nabla f\|\hi2\, d\mu,
    \end{align}
    for all smooth function $f: \real\hi d\to  \real$ such that the integration are finite.
\end{definition}
The log Sobolev inequality holds for any strongly log-concave measure on $\real\hi d$, that is, a measure of having density $e\hi{V(x)}$ and Hessian $\nabla\hi2 V\succeq \alpha I\lo d$ ($\alpha > 0$). The log Sobolev inequality indicates the following transport inequality: 
\begin{align*}
    W\lo2\hi2(\nu, \mu)\le \kappa\hi2\mathrm{KL}(\nu|\mu), \quad\text{for any}\,\, \nu\in \ca P(\real\hi d),
\end{align*}
where $\mathrm{KL}(\nu|\mu)$ is the Kullback–Leibler divergence. Please see \citet{gozlan2010transport} for more details on transport inequalities. For $j = 1,\dots, p$, we assume the following regularity conditions:
\begin{itemize}
    \item[(A3)] $P\lo j$ satisfies the log Sobolev inequality with constant $\kappa\hi 2$;
    \item[(A4)] There exists $\tau >  0$ such that $W\lo 2(P\lo j, Q) > \tau$.
\end{itemize}
Assumption (A4) requires the non-Gaussianity signal to be significant. We then establish a 0-concentration inequality for  $\|\hat S\lo j - S\lo j\|\lo{L\lo 1(\hat P\lo j )}$ in Lemma \ref{lemma: concentration_s_hat_s}. 
\begin{lemma}\label{lemma: concentration_s_hat_s}
For $j = 1,\dots, p$, assume $\psi\lo j, j=1, \dots, p$ satisfy assumption (A1) and (A2). With assumptions (A3) and (A4), we have
\begin{align*}
    \mathbb P\left(\frac{1}{n}\sum_{i = 1}^n \left\|\hat S\lo j(X\hi i\lo j)-S\lo j (X\hi i\lo j)\right\|\ge \varepsilon\right)\le  \exp\left\{-Cn\left[(\varepsilon - Cn\hi{-\frac{1}{4}})\lo+\right]\hi 2\right\}. 
\end{align*}
\end{lemma}

Similar to the CMC setting,  let $S\hii{w}\lo j(x)$ and $\hat S\hii{w}\lo j(x)$ be a winsorized version of $S\lo j(x)$ and $\hat S\lo j(x)$ with threshold $\sqrt{2\log n}$, respectively. The concentration inequality in Lemma \ref{lemma: concentration_s_hat_s} applies to $\hat S\hii{w}$ and $S\hii{w}$.
\begin{coro}\label{coro: concentration_win_s}
With same assumptions in Lemma \ref{lemma: concentration_s_hat_s}, we have
\begin{align*}
    \mathbb P\left(\frac{1}{n}\sum_{i = 1}^n \left\|\hat S\hii{w}\lo j(X\hi i\lo j)-S\hii{w} (X\hi i\lo j)\right\|\ge \varepsilon\right)\le  \exp\left\{-Cn\left[(\varepsilon - Cn\hi{-\frac{1}{4}})\lo+\right]\hi 2\right\}.
\end{align*}

\end{coro}
Compared with the concentration inequalities in Lemma \ref{lemma: concentration_t_hat_t}, the mean convergence rate is sharpened to $n\hi{-\frac{1}{4}}$ from $n\hi{-\frac{1}{4\lor d}}(\log n)\hi{\zeta\lo{d}}$. Using Lemma \ref{lemma: concentration_s_hat_s}, we can also obtain a sharper convergence rate for the covariance matrix estimation than Theorems \ref{thm: main} by replacing $n\hi{-\frac{1}{4\lor d}}(\log n)\hi{\zeta\lo{d}}$ with $n\hi{-\frac{1}{4}}$. When the principal non-Gaussian dimension $r > 1$, stronger assumptions are required to guarantee the consistency of the estimated subspace $\hat U\lo j$. We leave the investigation of such assumptions for future research.

\section{Simulation}\label{sec: simulation}
\subsection{Graph Learning with Two Attributes}\label{sec: simu_cmc}
We evaluate the numerical performance of thresholding, group graphical lasso, and neighborhood group lasso estimators (with ``and'' rule) for CMC-GGM when the dimension on each node is $2$. We also compare the performances of the three estimation methods with the Gaussian Graphical Model (GGM) and Copula Gaussian Graphical Model (Copula-GGM). To design the experiments, we first generate $d\times p$-dimensional Gaussian random vectors with mean $0$ and the following block precision matrix $\Theta$: 
\begin{itemize}
    \item[(A)] Banded precision matrix: For $j = 1,\dots, p$, $\Theta\lo{j,j} = I\lo d$; for $j \ge 2$, $\Theta\lo{j,j-1} = \Theta\lo{j-1,j} = 0.4I\lo d$; for $j \ge 3$, $\Theta\lo{j,j-2} = \Theta\lo{j-2, j} = 0.2I\lo d$.
    \item[(B)] Random precision matrix: Divide the graph into two connected parts, that is, let $\Theta = \diag(\Theta\hi 1, \Theta\hi 2)$, where $\Theta\hi \ell\in \real\hi{p/2\times p/2}$ for $\ell=1, 2$. For any $j\neq k, \ell=1, 2$, $\Theta\lo{j,k}\hi \ell = \xi I\lo d$, where $\xi = 0.3$ with probability $0.1$ and $0$ otherwise; $\Theta\lo{j, j}=\delta\cdot I\lo d$, where $\delta$ is chosen to guarantee the positive definiteness of $\Theta$.
    \item[(C)] Hub-connected precision matrix: Generate a graph’s edge set $E$ as follows. First, for all $j < k$, we set $E\lo{jk} = 1$ with probability $0.01$, and $0$ otherwise. Next, we randomly select $h=2$ hub nodes and set the elements of the corresponding rows and columns of $E$ equal $1$ with probability $0.5$ and $0$ otherwise. For $s = 1,\dots, M$, generate $p\times p$ matrix $\Omega\lo s$ by
    \begin{align*}
        {\Omega}\lo{s, jk} = {\Omega}\lo{s, kj} = \begin{cases}
        \delta, & \text{if}\,\, j=k,\\
        0, & \text{if}\,\, E\lo{jk} = 0,\\
        \xi, & \text{if}\,\, E\lo{jk} = 1,
        \end{cases}
    \end{align*}
    where $\xi\sim U([-0.75,-0.25]\cap[0.25,0.75])$ and $\delta$ is chosen to guarantee the positive definiteness of $\Omega\lo s$. Let $\Omega$ be the block diagonal matrix $\diag(\Omega\lo1,\dots, \Omega\lo d)$. The precision matrix is rearranged as $(\Theta \lo{jk})\lo{st} = (\Omega\lo{st})\lo{jk}$ for all $1\le j,k \le p$ and $1\le s,t\le d$.
\end{itemize}
\newcommand{\rulesep}{\unskip\ \vrule\ }
\begin{figure}[ht!]
     \centering
     \includegraphics[width=0.8\textwidth]{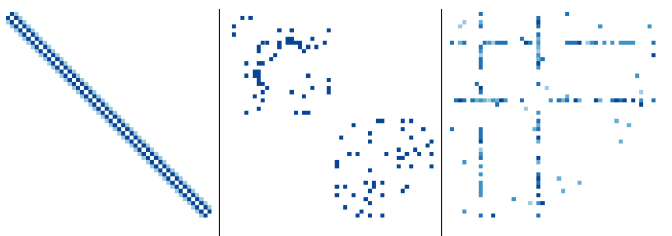}
      \caption{Visualization of graph structures for Models (A), (B), and (C), from left to right.}
        \label{fig: graph_map}
\end{figure} 
Figure \ref{fig: graph_map} shows the graph structure of Models (A), (B), and (C). Let $\Sigma = \Theta\inv$ be the covariance matrix, and $\diag(\Sigma)\hi{-1/2} = \diag(\Sigma\lo{11}\hi{-1/2}, \dots, \Sigma\lo{pp}\hi{-1/2})$ be the block diagonal matrix. We generate $Z = (Z\lo 1\trans, \dots, Z\lo p\trans)\trans \sim N(0,\diag(\Sigma)\hi{-1/2}\Theta\inv\diag(\Sigma)\hi{-1/2})$, which ensures that the distributions of the node vectors are standard bivariate Gaussian. We refer to $Z$ as the oracle Gaussian data that determines the graph structure. We then consider the following three transformations from $Z$ to the observed data $X$:
\begin{itemize}
    \item[(i)] $X = Z$, which correspnds to observing Gaussian data;
    \item[(ii)] $X\lo {js} = f\lo{js}(Z\lo {js})$, for $j = 1,\dots, p$ and $s=1,2$, where $f\lo {js}$ is the exponential transformation, defined by
    \begin{align*}
        f\lo {js}(z) = \sigma\lo {js}\left(\frac{\exp(z) - E[\exp(Z\lo {js})]}{\sqrt{\var(\exp(Z\lo {js}))}}\right),
    \end{align*}
    where $\sigma\lo{js}$ is the standard deviation of $Z\lo{js}$.
    \item[(iii)] For $j = 1,\dots, p$, define $f\lo j: \real\hi2\to\real\hi2$ as $f\lo j = (f\lo{j1}, f\lo{j2})$, where $f\lo{js}, s=1,2$ are the same as in (ii). Define $ g\lo j: \real\hi2\to\real\hi2$ as $g\lo j(z)= U (f\lo j(U\trans z))$, where $U = ((1,-1)\trans (1,1)\trans)/\sqrt{2}\in\real\hi{2\times 2}$. Let $X\lo j = g\lo j(Z\lo j)$. 
\end{itemize}
We note that transformation (ii) is coordinatewise monotonic, which guarantees that the subvector $X\lo j$ follows a nonparanormal distribution but not a multivariate Gaussian. We design the marginal transformation $f\lo{js}$ to preserve the marginal standard deviation of the transformed data. Transformation (iii) is not coordinate-wise monotonic but cyclically monotonic, which ensures that $X$ has a CMCG distribution. Similar models with transformation (iii) are considered in \cite{bryan2021multirank} for semiparametric CCA modeling.
\begin{table}[ht]
\centering
\resizebox{\columnwidth}{!}{
\begin{tabular}{cllllllllll}
 \hline
&   \multicolumn{3}{c}{group-glasso}  & \multicolumn{3}{c}{thresholding}  &  \multicolumn{3}{c}{nbd-group-lasso}  \\
\cline{2-10}
Models& CMC & Copula & Linear & CMC & Copula & Linear & CMC & Copula & Linear \\
\hline
\multirow{2}{*}{A-i}  
  & 0.977 & 0.982 & 0.983 & 0.976 & 0.982 & 0.983 & 0.991 & 0.993 & \textbf{0.994} \\ 
  & (0.004) & (0.003) & (0.003) & (0.007) & (0.005) & (0.005) & (0.002) & (0.002) & (0.001) \\ 
\hline
\multirow{2}{*}{A-ii} 
  & 0.974 & 0.982 & 0.862 & 0.974 & 0.982 & 0.833 & 0.989 & \textbf{0.993} & 0.864 \\ 
  & (0.004) & (0.003) & (0.011) & (0.006) & (0.005) & (0.014) & (0.003) & (0.002) & (0.013) \\ 
\hline
\multirow{2}{*}{A-iii} 
  & 0.974 & 0.950 & 0.863 & 0.971 & 0.940 & 0.834 & \textbf{0.989} & 0.966 & 0.862 \\ 
  & (0.004) & (0.007) & (0.011) & (0.006) & (0.008) & (0.013) & (0.003) & (0.005) & (0.011) \\ 
\hline
\end{tabular}
}
\caption{Means and standard errors (in parentheses) for AUC of Model A}
\label{table: mggm_model_a}
\end{table}

\begin{table}[ht]
\centering
\resizebox{\columnwidth}{!}{
\begin{tabular}{cllllllllll}
 \hline
&   \multicolumn{3}{c}{group-glasso}  & \multicolumn{3}{c}{thresholding}  &  \multicolumn{3}{c}{nbd-group-lasso}  \\
\cline{2-10}
Models&  CMC & Copula & Linear & CMC & Copula & Linear & CMC & Copula & Linear \\
\hline
\multirow{2}{*}{B-i}  
 & 0.928 & 0.940 & \textbf{0.942} & 0.777 & 0.792 & 0.797 & 0.918 & 0.929 & 0.931 \\ 
 & (0.012) & (0.011) & (0.011) & (0.017) & (0.017) & (0.017) & (0.013) & (0.013) & (0.013) \\ 
\hline
\multirow{2}{*}{B-ii}
 & 0.922 & \textbf{0.939} & 0.776 & 0.77 & 0.791 & 0.613 & 0.914 & 0.929 & 0.753 \\ 
 & (0.016) & (0.014) & (0.02) & (0.02) & (0.021) & (0.023) & (0.013) & (0.013) & (0.018) \\ 
\hline
\multirow{2}{*}{B-iii}
& \textbf{0.924} & 0.894 & 0.774 & 0.879 & 0.838 & 0.691 & 0.914 & 0.879 & 0.748 \\ 
 & (0.013) & (0.014) & (0.016) & (0.016) & (0.016) & (0.018) & (0.014) & (0.014) & (0.017) \\
\hline
\end{tabular}
}
\caption{Means and standard errors (in parentheses) for AUC of Model B}
\label{table: mggm_model_b}
\end{table}

\begin{table}[ht]
\centering
\resizebox{\columnwidth}{!}{
\begin{tabular}{cllllllllll}
 \hline
&    \multicolumn{3}{c}{group-glasso}  & \multicolumn{3}{c}{thresholding}  &  \multicolumn{3}{c}{nbd-group-lasso}  \\
\cline{2-10}
Models&  CMC & Copula & Linear & CMC & Copula & Linear & CMC & Copula & Linear \\
\hline
\multirow{2}{*}{C-i}  
  & 0.885 & 0.896 & \textbf{0.899} & 0.775 & 0.791 & 0.794 & 0.818 & 0.826 & 0.829 \\ 
  & (0.019) & (0.017) & (0.017) & (0.025) & (0.025) & (0.026) & (0.023) & (0.024) & (0.024) \\ 
\hline
\multirow{2}{*}{C-ii}    
  & 0.880 & \textbf{0.896} & 0.746 & 0.767 & 0.791 & 0.638 & 0.812 & 0.826 & 0.687 \\ 
  & (0.018) & (0.017) & (0.028) & (0.025) & (0.025) & (0.038) & (0.023) & (0.024) & (0.029) \\ 
 \hline
\multirow{2}{*}{C-iii}  
  & \textbf{0.877} & 0.849 & 0.742 & 0.760 & 0.728 & 0.637 & 0.815 & 0.784 & 0.687 \\ 
  & (0.019) & (0.020) & (0.028) & (0.037) & (0.032) & (0.037) & (0.023) & (0.023) & (0.030) \\
\hline
\end{tabular}
}
\caption{Means and standard errors (in parentheses) for AUC of Model C}
\label{table: mggm_model_c}
\end{table}

We conducted simulations with a network size of $p=100$ and sample size of $n=300$, repeating each simulation $50$ times. We calculated the true positive rate (TPR) and false positive rate (FPR) for $20$ different threshold values $\epsilon\lo n$ or tuning parameters $\lambda\lo n$, with which we generated a receiver operating characteristic (ROC) curve.  The means and standard deviations (in parentheses) of the associated area-under-curve values (AUC) are reported in Tables \ref{table: mggm_model_a}, \ref{table: mggm_model_b}, and \ref{table: mggm_model_c}.

Our results indicate the following points: (a) when the data are generated from a Gaussian distribution (transformation (i)), linear transformation gives the most accurate estimation, but the difference among the three transformation methods is not significant; (b) when the data are generated from a copula Gaussian distribution (transformation (ii)), CMC-GGM and Copula-GGM give provide much more accurate estimation than GGM; (c) when the data are generated from a CMCG distribution, CMC-GGM significantly improves estimation accurate from Copula-GGM; (d) for Model A, the neighborhood group lasso provides higher AUC scores than the group glasso and thresholding; and (e) for the more challenging Models B and C, the group graphical lasso outperforms the neighborhood group lasso while thresholding performs poorly.

\begin{figure}[ht!]
    \centering
    \includegraphics[width=\textwidth]{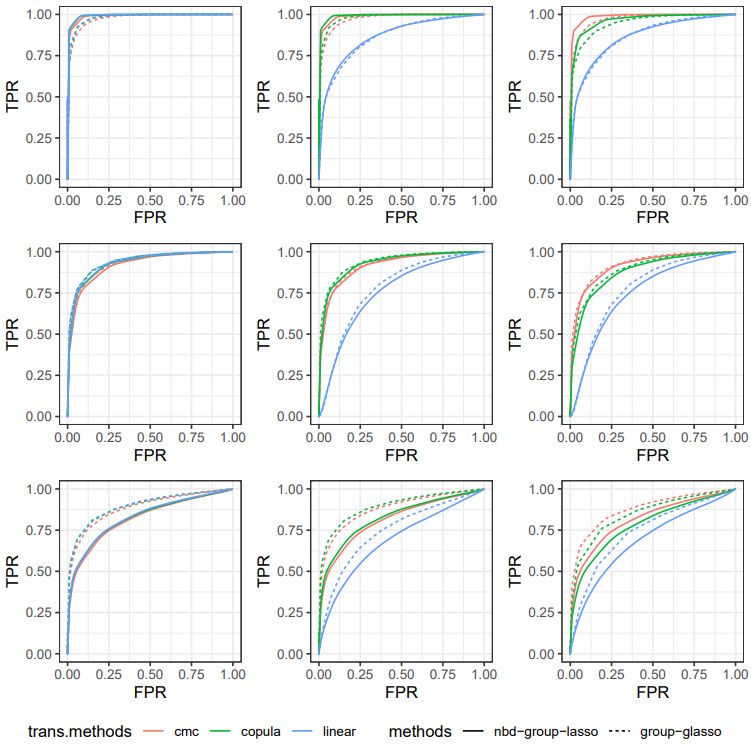}
    \caption{Comparison of average estimated ROC curves over $50$ simulation runs: Model A-i, A-ii and A-iii (top row); Model B-i, B-ii, and B-iii (middle row); Model C-i, C-ii, and C-iii (bottom row).}
    \label{fig:roc_p100}
\end{figure}
To visualize the comparison, in Figure \ref{fig:roc_p100} we display the average ROC curves of the methods estimated by graphical glasso and neighborhood group lasso. From the plot, we are further convinced that the CMC transformation methods (red curves) have the best overall performance when the data are generated from non-copula Gaussian distribution (right column).

\begin{figure}[ht!]
    \centering
    \includegraphics[width=\textwidth]{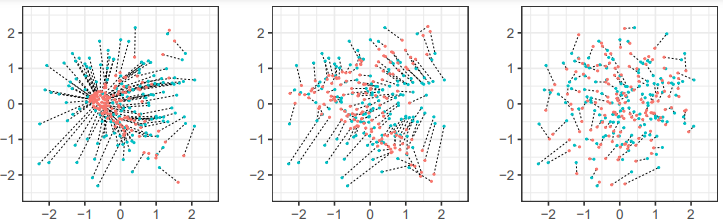}
    \caption{Visualization of transformations for setting (iii): green points are oracle data, red points are data from (Left) observations; (Middle) copula transformation; (Right) CMC transformation.}
    \label{fig: transform}
\end{figure}

To better illustrate why copula models fail under setting (iii), we show the copula and CMC transformation results in Figure \ref{fig: transform}. We generate $n=300$ samples from Model A-iii on one node. The green points are the oracle Gaussian data that we want to recover and use to estimate the precision matrix further. From left to right, the red points are observations, copula-transformed data, and CMC-transformed data, respectively. We see that in the middle panel, after the copula transformation, the red points are marginal Gaussian on both coordinates but jointly show a triangular shape, which deviates from the joint Gaussian distribution. Consequently, the values on both coordinates are estimated poorly, affecting the further estimation of the precision matrix. In contrast, the CMC transformation can better recover the underlying oracle points, as demonstrated in the right panel.

\subsection{Graph Learning with More Than Two Attributes}\label{sec: simu_pcmc}
In this subsection, we compare the performance of Copula-GGM, CMC-GGM, and PCMC-GGM when the number of attributes on each node is larger than $2$. We first generate Gaussian data from Model (A) in Section \ref{sec: simu_cmc}. For $j=1, \dots, p$, define $f\lo j: \real\hi {r} \to\real\hi {r}$ by $f\lo j = (f\lo{j1}, \cdots, f\lo{jr})$, where $f\lo {js}$ is univariate monotonic function. We consider two choices for $f\lo {js}$: the first one is the exponential map defined in Section \ref{sec: simu_cmc} (ii); the second one is the cubic map defined by
\begin{align*}
    f\lo{js}(z) = \sigma\lo{js}\left(\frac{z\hi3 - E[Z\lo{js}\hi3]}{\sqrt{\var(Z\lo{js}\hi3)}}\right),
\end{align*}
where $\sigma\lo{js}$ is the standard deviation of $Z\lo{js}$. Then we define $g\lo j:\real\hi{d}\to \real\hi d$ as $g\lo{j}(z) = U(f\lo j(U\trans z)) + VV\trans z$, where $U\in\real\hi{d\times r}$ and $(U, V)\in \real\hi {d\times d}$ is a orthogonal matrix. Let $X\lo j = g\lo j(Z\lo j)$. Under this setting, the non-Gaussian observations $X\lo j$ and the oracle Gaussian data $Z\lo j$ only differ on a $r$-dimensional subspace spanned by $U$. To be consistent with the notations in Section \ref{sec: simu_cmc}, we call the models A-iv-exp and A-iv-cubic. Here, we use $n =300, p = 50$, and $d\in\{3, 5, 7\}$. We let $r=1$ and $U = (1, 1, 0,\dots, 0)\trans/\sqrt{2}$, that is, there is one non-Gaussian direction contained in $X\lo j$. More numerical results on $r> 1$ can be found in the Supplementary Material.

\begin{figure}[ht!]
    \centering
    \includegraphics[width=\textwidth]{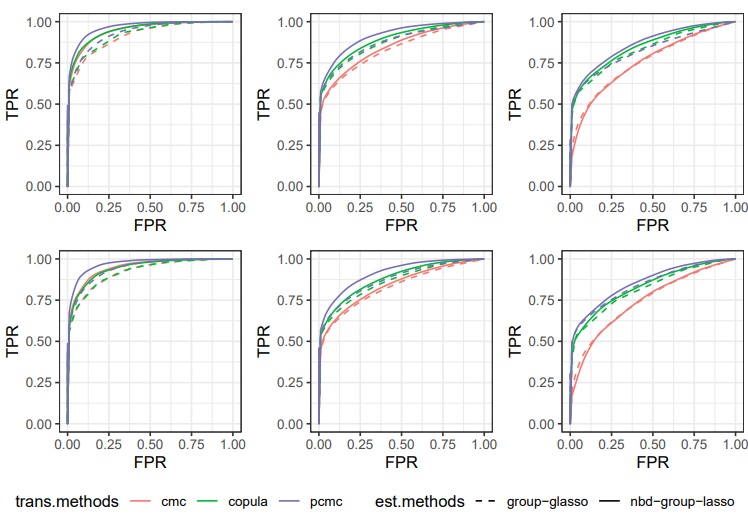}
    \caption{Comparison of average estimated ROC curves over $50$ simulation runs among PCMC-GGM, CMC-GGM, and Copula-GGM; First row:  Model A-iv-exp; Second row: Model A-iv-cubic; Columns from left to right corresponds to $d= 3$, $5$, and $7$.}
    \label{fig:roc_proj_r1}
\end{figure}

For PCMC-GGM, we independently generate $10$ samples and use the Riemannian block coordinate descent (RBCD) algorithm proposed in \citet{huang2021riemannian} to solve $\hat U\lo j\hi l, j = 1,\dots, p, l = 1,\dots, 10$. We then take the extrinsic average of 10 repeats to obtain the estimator $\hat U\lo j$, which is the first eigenvector of $\sum\lo{l=1}\hi{10} \hat U\lo j\hi l (\hat U\lo j\hi l)\trans$. We assume that the true dimension $1$ is known. Determining the dimension for the projected subspace distance in a more systematic way is beyond the scope of this paper. We then solve the OT problem between projected samples $\{\hat U\lo j X\lo j\hi i\}\lo{i=1}\hi n$ and $\{\hat U\lo jZ\lo j\hi i\}\lo{i=1}\hi n$ for $j =1, \dots, p$ to get an estimate optimal transport $\hat T\lo j$. The PCMC transformation is then estimated by \eqref{eq: hat_S}. After 50 repeats of the experiments, we plot the average ROC curves of the PCMC-GGM, CMC-GGM, and Copula-GGM in Figure \ref{fig:roc_proj_r1}. We also report the means and standard deviations (in parentheses) of the associated AUC scores in Table \ref{table: proj_r1}. We observe from the table that the accuracy of CMC-GGM drops sharply with the increase of the attribute dimension. Overall, PCMC-GGM outperforms Copula-GGM and CMC-GGM. 

\begin{table}[ht]
\centering
\resizebox{\columnwidth}{!}{
\begin{tabular}{clllllllllll}
  \hline
&   &  \multicolumn{3}{c}{group-glasso} & \multicolumn{3}{c}{nbd-group-lasso} & \multicolumn{3}{c}{thresholding}  \\
\cline{3-11}
Model & $d$  & CMC & Copula & PCMC & CMC & Copula & PCMC & CMC & Copula & PCMC \\
\hline
  \multirow{6}{*}{A-iv-exp}  
   & 3   & 0.921 & 0.920 & 0.942 & 0.949 & 0.944 & \textbf{0.970} & 0.917 & 0.915 & 0.949 \\ 
   &     & (0.013) & (0.014) & (0.01) & (0.009) & (0.011) & (0.007) & (0.017) & (0.014) & (0.013) \\ 
   & 5   & 0.827 & 0.863 & 0.875 & 0.840 & 0.882 & \textbf{0.915} & 0.662 & 0.744 & 0.789 \\ 
   &     & (0.018) & (0.014) & (0.052) & (0.018) & (0.014) & (0.012) & (0.028) & (0.030) & (0.026) \\ 
   & 7   & 0.747 & 0.820 & 0.845 & 0.744 & 0.834 & \textbf{0.860} & 0.593 & 0.696 & 0.739 \\ 
   \cline{2-11}
 \multirow{6}{*}{A-iv-cubic}
   &     & (0.019) & (0.016) & (0.018) & (0.019) & (0.017) & (0.015) & (0.031) & (0.024) & (0.020) \\ 
   & 3   & 0.914 & 0.922 & 0.933 & 0.948 & 0.951 & \textbf{0.967} & 0.918 & 0.914 & 0.94 \\ 
   &     & (0.011) & (0.013) & (0.013) & (0.009) & (0.011) & (0.008) & (0.013) & (0.019) & (0.012) \\ 
   & 5   & 0.83 & 0.874 & 0.877 & 0.845 & 0.892 & \textbf{0.918} & 0.677 & 0.76 & 0.788 \\ 
   &     & (0.019) & (0.017) & (0.027) & (0.018) & (0.015) & (0.014) & (0.028) & (0.020) & (0.022) \\ 
   & 7   & 0.758 & 0.833 & 0.826 & 0.755 & 0.849 & \textbf{0.867} & 0.602 & 0.715 & 0.739 \\ 
   &     & (0.022) & (0.020) & (0.050) & (0.020) & (0.019) & (0.016) & (0.028) & (0.029) & (0.028) \\ 
   \hline
\end{tabular}
}
\caption{Means and standard errors (in parentheses) for AUC for Models A-iv-exp and A-iv-cubic}\label{table: proj_r1}
\end{table} 

\begin{figure}[ht!]
    \centering
    \includegraphics[width=\textwidth]{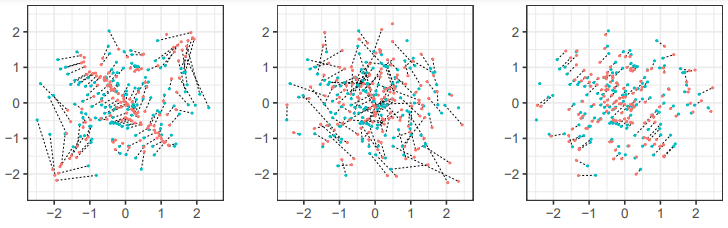}
    \caption{Visualization of transformations in Model A-iv-cubic: red points are oracle data, green points are data from (L) copula transformation; (M) CMC transformation; (R) PCMC transformation.}
    \label{fig: transform_pcmc_r1}
\end{figure}

We provide a visualization of the Copula, CMC, and PCMC transformations in Figure \ref{fig: transform_pcmc_r1}. Specifically, we consider the data on one node in Model A-iv-cubic when $d = 5$ and plot the data on the first two coordinates. The green points are the oracle Gaussian data. From left to right, the red points are the transformed data from Copula, CMC, and PCMC transformations, respectively. We observe that in the left panel, after the copula transformation, red points still show non-Gaussianity along direction $(1,1,0,\dots, 0)\trans$. In the middle panel, after the CMC transformation, red points can not recover green points accurately due to the large estimation error arising from solving a high-dimensional OT problem. In contrast, the PCMC transformation can accurately recover the underlying oracle points, as shown in the right panel.

\section{Real Applications}\label{sec: application}
In this section, we consider two data applications where multi-attributes occur naturally. Since the ground truth is unknown in the data application, the goal of the analysis is only to visualize and explore the underlying dependency structures of the data. We will focus on showing the non-Guassianality of the data and compare the performance of GGM, Copula-GGM, and CMC-GGM. We use the group glasso to estimate the graph, which performs the best in the simulations.

\subsection{Gene/Protein Regulatory Network Inference}
We applied the CMC-GGM to reconstruct networks on breast cancer data sets from the National Cancer Institute   (\url{https://www.cancer.gov/}), referred to as NCI-60 and analyzed in \citet{katenka2011multi,kolar2014graph} and \citet{chiquet2019multiattribute}. The data set contains protein profiles (reverse-phase lysate arrays for 92 antibodies) and gene profiles (normalized RNA microarray intensities from Human Genome U95 Affymetrix chip-set for about 9000 genes). \citet{katenka2011multi} constructed a `concensus' data set containing 91 ($p = 91$) protein/gene profiles matched in pairs ($d = 2$) by common Entrez identifiers across 60 ($n=60$) cancer cells.

\begin{table}[ht!]
    \centering
    \begin{tabular}{l c cc}
    \hline
         & GGM &Copula-GGM & CMC-GGM \\
         \hline
         Number of edges&  299 & 309 & 168\\
         Number of shared edges with GGM & ** & 183 & 129\\
         Avg Node Degree & 6.571 & 6.791 & 3.692\\
         Number of nodes with non-Gaussianity & 69 & 9 & 0\\
    \hline     
    \end{tabular}
    \caption{Summary statistics for gene/protein networks estimated by GGM, Copula-GGM, and CMC-GGM}
    \label{table: gene_protein_summary}
\end{table}

\begin{figure}[ht!]
    \centering
    \begin{subfigure}[b]{0.35\textwidth}
         \centering
         \includegraphics[width=\textwidth]{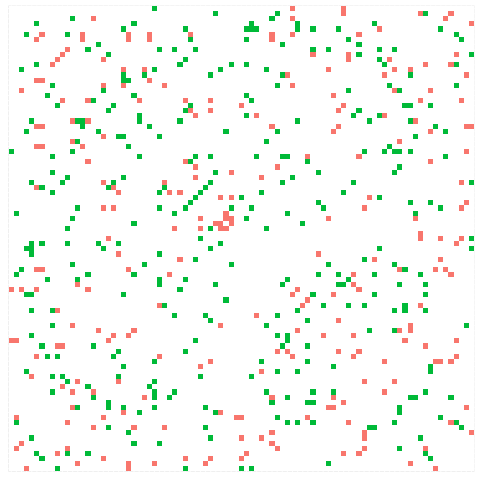}
         \caption{GGM; sparsity $0.073$.}
     \end{subfigure}
     \hspace{0.2in}
    \begin{subfigure}[b]{0.35\textwidth}
          \centering
          \includegraphics[width=\textwidth]{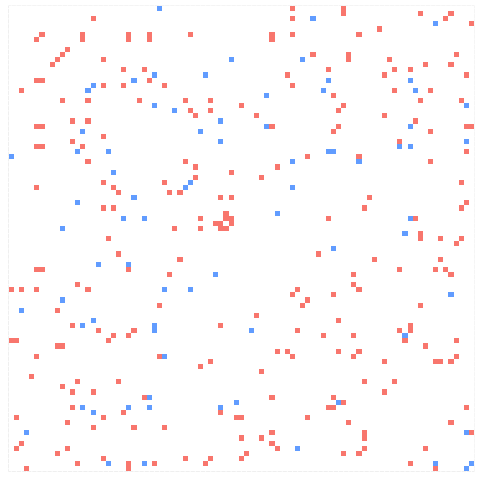}
          \caption{CMC-GGM; sparsity $0.053$.}
      \end{subfigure}
    \caption{Gene/Protein network; Red cells represent edges shared by (a) and (b); green cells represent edges unique to (a); and blue cells represent edges unique to (b).}
    \label{fig: gene_protein}
\end{figure}

We begin by applying the group glasso for GGM to fit the gene/protein network  and then compare it with the network constructed using the Copula-GGM and CMC-GGM. For each method, we use the 10-fold cross-validation to select the tuning parameters, resulting in dense networks with a sparsity of about $0.4$. We then estimate the network 50 times based on bootstrap samples. The stable graph was then constructed from the edges that appeared in at least $90\%$ of the bootstrap replications. Table \ref{table: gene_protein_summary} provides a few summary statistics for the estimated networks. To test the joint Gaussianity of the data on each node, we use the energy statistics \citep{szekely2013energy} with a permutation test of 499 replications. A node is identified to have non-Gaussian data if the p-value is less than $0.05$. Among all 91 nodes, 69 have non-Gaussian data, and 9 still exhibit non-Gaussianity even after copula transformations. Figure \ref{fig: gene_protein} shows the graph fitted by GGM and CMC-GGM, where red cells represent the edge shared by the two methods, green cells represent the edges unique to GGM, and blue cells represent the edges unique to CMC-GGM. The differences in networks require a closer biological inspection based on domain knowledge.

\subsection{Color Texture Images}
\citet{pavez2016generalized} and \citet{pavez2018learning} built undirected graphs for grayscale images to infer the dependence of a pixel on neighboring pixels. \citet{tugnait2021sparse} extended this idea to build a multi-attribute graphical model for colored images, where each node represents three attributes (RGB components) from a pixel.

\begin{figure}[ht!]
    \centering
    \begin{subfigure}[b]{0.25\textwidth}
         \centering
         \includegraphics[width=\textwidth]{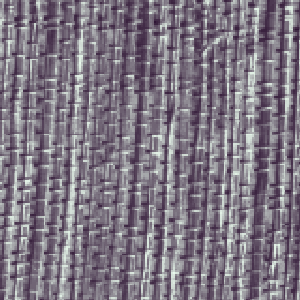}
         \caption{image 1}
     \end{subfigure}
     \hspace{0.1in}
      \begin{subfigure}[b]{0.35\textwidth}
          \centering
          \includegraphics[width=\textwidth]{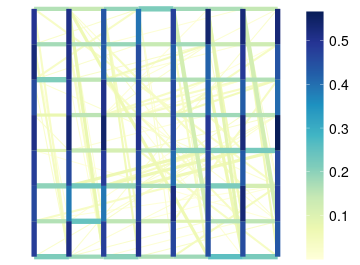}
          \caption{GGM; sparsity $0.174$.}
      \end{subfigure}
      \begin{subfigure}[b]{0.35\textwidth}
          \centering
          \includegraphics[width=\textwidth]{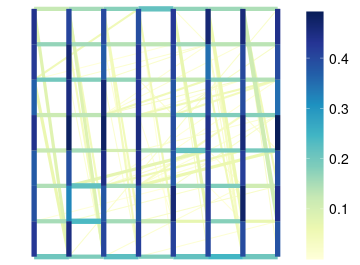}
          \caption{CMC-GGM; sparsity $0.160$.}
      \end{subfigure}
      \begin{subfigure}[b]{0.25\textwidth}
         \centering
         \includegraphics[width=\textwidth]{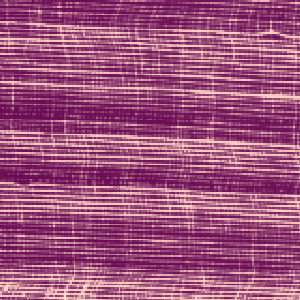}
         \caption{image 2}
     \end{subfigure}
     \hspace{0.1in}
      \begin{subfigure}[b]{0.35\textwidth}
          \centering
          \includegraphics[width=\textwidth]{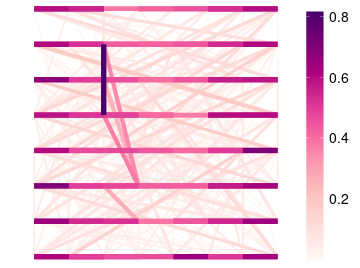}
          \caption{GGM; sparsity $0.203$.}
      \end{subfigure}
      \begin{subfigure}[b]{0.35\textwidth}
          \centering
          \includegraphics[width=\textwidth]{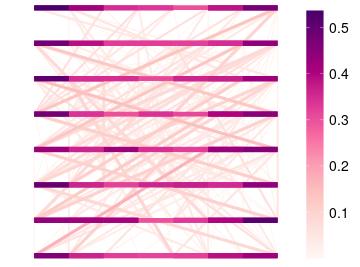}
          \caption{CMC-GGM; sparsity $0.187$.}
      \end{subfigure}
    \caption{Color texture graph}
    \label{fig: texture_plot}
\end{figure}

To evaluate our method, we selected two images (Image 79 and Image 105) from the Colored Brodatz Texture Database (\url{https://multibandtexture.recherche.usherbrooke.ca/}). We label image 79 as image 1 and image 105 as Image 2. Two images are read as a raster object with dimension $640 \times 640 \times 3$ in R. For image 1, we extracted rows 481 through 640 and columns 481 through 640. For image 2, we extracted rows 321 through 480 and columns 1 through 160. This creates the $160 \times 160$ patches used to build image
graphs, visualized in Figure \ref{fig: texture_plot}(a) and (d). The patches were then partitioned into non-overlapping $8 \times 8$ blocks and then vectorized into 64-dimensional pixel vectors. For each node, we have three attributes associated with RGB decompositions. Therefore, we have a data set with sizes $n=400$, $p=64$, and $d=3$. We found that the data showed non-Gaussianity with p-values less than $0.05$ from an energy test at all nodes. Even after copula transformations, non-Gaussianity still persisted. We compared the performance of GGM and CMC-GGM on this data set. We used the BIC-based method to select the tuning parameters.  For image 1, CMC-GGM and GGM obtained graphs with sparsities of $0.162$ and $0.174$, respectively. For image 2, CMC-GGM and GGM drive graphs with sparsities of $0.183$ and $0.211$, respectively. 

The estimated graphs are visualized in Figure \ref{fig: texture_plot}. The color and width of the links indicate the edge weights, which are taken as the Frobenius norm of the block matrix $\|\hat\Theta\lo{jk}\|\lo F$. By comparing the estimated graphs with the raw texture images, we can observe that strong links capture the principle texture orientations. Specifically, for image 1, vertical and some horizontal directions are the primary orientation of the texture; for image 2, the horizontal direction is the primary orientation. By comparing the graph (b) and (d), and (e) and (f), we see that using CMC transformation helps to estimate more precise graphs that match the raw image. On the other hand, the GGM may retain more error signals, such as the vertical signals on the upper left corner of graph (e).

\section{Conclusion}

This paper presents a new semiparametric copula graphical model for multi-attribute data based on the newly introduced cyclically monotone copula. The proposed model is more flexible than the existing copula Gaussian graphical model that only performs coordinatewise Gaussianization. We demonstrate both theoretical and numerical properties of the proposed methods. In future work, it will be interesting to study other types of graphical models such as \cite{xue2012nonconcave,yang2015graphical,tao2021additive,lee2022estimating} using optimal transport theory.

{
\bibliographystyle{agsm}
\bibliography{ref}
}

\end{document}